\documentclass{article}

\usepackage{microtype}
\usepackage{graphicx}
\usepackage{subfigure}
\usepackage{booktabs} 

\usepackage{hyperref}

\usepackage[accepted]{icml2020}

\usepackage{graphicx}
\usepackage{subfigure}
\usepackage{amsmath, amsfonts, bm}
\usepackage{amsthm}
\usepackage{multirow}

\newtheorem{prop}{Proposition}

\newtheorem{definition}{Definition}

\newcommand{\thetav}{\boldsymbol \theta}
\newcommand{\Thetav}{\boldsymbol \Theta}

\newcommand{\zerov}{\boldsymbol 0}

\newcommand{\gv}{\mathbf{g}}

\newcommand{\wv}{\mathbf{w}}
\newcommand{\xv}{\mathbf{x}}
\newcommand{\yv}{\mathbf{y}}

\newcommand{\Iv}{\mathbf{I}}

\newcommand{\argmax}{\operatornamewithlimits{argmax}}
\newcommand{\argmin}{\operatornamewithlimits{argmin}}
\newcommand{\argTopk}{\operatornamewithlimits{argTopk}}
\newcommand{\argTopm}{\operatornamewithlimits{argTopm}}
\renewcommand{\algorithmicrequire}{\textbf{Input:}}
\renewcommand{\algorithmicensure}{\textbf{Output:}}

\icmltitlerunning{Learning Optimal Tree Models under Beam Search}

\begin{document}

\twocolumn[
\icmltitle{Learning Optimal Tree Models under Beam Search}



\icmlsetsymbol{equal}{*}

\begin{icmlauthorlist}
\icmlauthor{Jingwei Zhuo}{to}
\icmlauthor{Ziru Xu}{to}
\icmlauthor{Wei Dai}{to}
\icmlauthor{Han Zhu}{to}
\icmlauthor{Han Li}{to}
\icmlauthor{Jian Xu}{to}
\icmlauthor{Kun Gai}{to}
\end{icmlauthorlist}

\icmlaffiliation{to}{Alibaba Group}
\icmlcorrespondingauthor{Jingwei Zhuo}{zjw169463@alibaba-inc.com}

\icmlkeywords{Machine Learning, ICML}

\vskip 0.3in
]



\printAffiliationsAndNotice{}  

\begin{abstract}
Retrieving relevant targets from an extremely large target set under computational limits is a common challenge for information retrieval and recommendation systems. 
Tree models, which formulate targets as leaves of a tree with trainable node-wise scorers, have attracted a lot of interests in tackling this challenge due to their logarithmic computational complexity in both training and testing.
Tree-based deep models (TDMs) and probabilistic label trees (PLTs) are two representative kinds of them. 
Though achieving many practical successes, existing tree models suffer from the training-testing discrepancy, where the retrieval performance deterioration caused by beam search in testing is not considered in training.
This leads to an intrinsic gap between the most relevant targets and those retrieved by beam search with even the optimally trained node-wise scorers.
We take a first step towards understanding and analyzing this problem theoretically, and develop the concept of \emph{Bayes optimality under beam search} and \emph{calibration under beam search} as general analyzing tools for this purpose.
Moreover, to eliminate the discrepancy, we propose a novel algorithm for learning optimal tree models under beam search. 
Experiments on both synthetic and real data verify the rationality of our theoretical analysis and demonstrate the superiority of our algorithm compared to state-of-the-art methods.
\end{abstract}

\section{Introduction}\label{sec:intro}
Extremely large-scale retrieval problems prevail in modern industrial applications of information retrieval and recommendation systems.
For example, in online advertising systems, several advertisements need to be retrieved from a target set containing tens of millions of advertisements and presented to a user in tens of milliseconds. 
The limits of computational resources and response time make models, whose computational complexity scales linearly with the size of target set, become unacceptable in practice.

Tree models are of special interest to solve these problems because of their ability in achieving logarithmic complexity in both training and testing.
Tree-based deep models (TDMs) \cite{zhu2018learning,zhu2019joint,you2019attentionxml} and Probabilistic label trees (PLTs) \cite{jasinska2016extreme,prabhu2018parabel,wydmuch2018no} are two representative kinds of tree models.
These models introduce a tree hierarchy in which each leaf node corresponds to a target and each non-leaf node defines a pseudo target for measuring the existence of relevant targets on the subtree rooted at it.
Each node is also associated with a node-wise scorer which is trained to estimate the probability that the corresponding (pseudo) target is relevant.
To achieve logarithmic training complexity, a subsampling method is leveraged to select logarithmic number of nodes on which the scorers are trained for each training instance.
In testing, beam search is usually used to retrieve relevant targets in logarithmic complexity.

As a greedy method, beam search only expands parts of nodes with larger scores while pruning other nodes.
This character achieves logarithmic computational complexity but may result in deteriorating retrieval performance if ancestor nodes of the most relevant targets are pruned.
An ideal tree model should guarantee no performance deterioration when its node-wise scorers are leveraged for beam search.
However, existing tree models ignore this and treat training as a separated task to testing:
(1) Node-wise scorers are trained as probability estimators of pseudo targets which are not designed for optimal retrieval;
(2) They are also trained on subsampled nodes which are different to those queried by beam search in testing.
Such discrepancy makes even the optimal node-wise scorers w.r.t. training loss can lead to suboptimal retrieval results when they are used in testing to retrieve relevant targets via beam search.
To the best of our knowledge, there is little work discussing this problem either theoretically or experimentally. 

We take a first step towards understanding and resolving the training-testing discrepancy on tree models.
To analyze this formally, we develop the concept of \emph{Bayes optimality under beam search} and \emph{calibration under beam search} as the optimality measure of tree models and corresponding training loss, respectively.
Both of them serve as general analyzing tools for tree models.
Based on these concepts, we show that neither TDMs nor PLTs are optimal, and derive a sufficient condition for the existence of optimal tree models as well.
We also propose a novel algorithm for learning such an optimal tree model.
Our algorithm consists of a beam search aware subsampling method and an optimal retrieval based definition of pseudo targets, both of which resolve the training-testing discrepancy.
Experiments on synthetic and real data not only verify the rationality of our newly proposed concepts in measuring the optimality of tree models, but also demonstrate the superiority of our algorithm compared to existing state-of-the-art methods.

\section{Related Work}
{\bf Tree Models:} 
Research on tree models\footnote{There also exist models which usually build an ensemble of decision trees over instances instead of targets \cite{prabhu2014fastxml,jain2016extreme}. They are less relevant to our main focus.} has mainly focused on formulating node-wise scorers and the tree structure. 
For node-wise scorers, linear models are widely adopted \cite{jasinska2016extreme,wydmuch2018no,prabhu2018parabel}, while deep models \cite{zhu2018learning,zhu2019joint,you2019attentionxml} become popular recently.
For the tree structure, apart from the random tree \cite{jasinska2016extreme}, recent works propose to learn it either via hierarchical clustering over targets \cite{wydmuch2018no,prabhu2018parabel,khandagale2019bonsai} or under a joint optimization framework with node-wise scorers \cite{zhu2019joint}.
Without dependence on specific formulations of node-wise scorers or the tree structure, our theoretical findings and proposed training algorithm are general and applicable to these advances.

{\bf Bayes Optimality and Calibration:}
Bayes optimality and calibration have been extensively investigated on flat models \cite{lapin2017analysis,menon2019multilabel,yang2019consistency}, and they have also been used to measure the performance of tree models on hierarchical probability estimation \cite{wydmuch2018no}.
However, there is a gap between the performance on hierarchical probability estimation and that on retrieving relevant targets, since the former ignores beam search and corresponding performance deterioration. 
As a result, how to measure the retrieval performance of tree models formally remains an open question.
We fill this void by developing the concept of Bayes optimality under beam search and calibration under beam search.

{\bf Beam Search in Training:} 
Formulating beam search into training to resolve the training-testing discrepancy is not a new idea.
It has been extensively investigated on structured prediction models for problems like machine translation and speech recognition \cite{daume2005learning,xu2007learning,ross2011reduction,wiseman2016sequence,goyal2018continuous,negrinho2018learning}. 
Though performance deterioration caused by beam search has been analyzed empirically \cite{cohen2019empirical}, it still lacks a theoretical understanding.
Besides, little effort has been made to understand and resolve the training-testing discrepancy on tree models.
We take a first step towards studying these problems both theoretically and experimentally.

\section{Preliminaries}\label{sec:preliminary}


\subsection{Problem Definition}

Suppose $\mathcal{I}=\{1,...,M\}$ with $M \gg 1$ is a target set and $\mathcal{X}$ is an observation space, we denote an instance\footnote{This summarizes many practical applications. For example, in recommendation systems, an instance corresponds to an interaction between users and items, where $\xv$ denotes the user information and $\mathcal{I}_{\xv}$ denotes the items in which the user are interested.} as $(\xv,\mathcal{I}_{\xv})$, implying that an observation $\xv \in \mathcal{X}$ is associated with a subset of relevant targets $\mathcal{I}_{\xv} \subset \mathcal{I}$, which usually satisfies $|\mathcal{I}_{\xv}| \ll M$.
For notation simplicity, we introduce a binary vector $\yv \in \mathcal{Y}=\{0,1\}^M$ as an alternative representation for $\mathcal{I}_{\xv}$, where $y_j=1$ implies $j \in \mathcal{I}_{\xv}$ and vice versa.
As a result, an instance can also be denoted as $(\xv,\yv) \in \mathcal{X} \times \mathcal{Y}$.

Let $p:\mathcal{X} \times \mathcal{Y} \to \mathbb{R}^+$ be a probability density function for data which is unknown in practice, we slightly abuse notations by regarding an instance $(\xv,\yv)$ as either the random variable pair w.r.t. $p(\xv,\yv)$ or a sample of $p(\xv,\yv)$.
We also assume the training dataset $\mathcal{D}_{tr}$ and the testing dataset $\mathcal{D}_{te}$ to be the sets containing i.i.d. samples of $p(\xv,\yv)$. 
Since $\yv$ is a binary vector, we use the simplified notation $\eta_j(\xv)=p(y_j=1|\xv)$ for any $j \in \mathcal{I}$ in the rest of this paper.

Given these notations, the extremely large-scale retrieval problem is defined as to learn a model $\mathcal{M}$ such that its retrieved subset for any $\xv \sim p(\xv)$, denoted by either $\hat{\mathcal{I}}_{\xv}$ or $\hat{\yv}$, is as close as $\yv \sim p(\yv|\xv)$ according to some performance metrics. 
Since $p(\xv,\yv)$ is unknown in practice, such a model is usually learnt as an estimator of $p(\yv|\xv)$ on $\mathcal{D}_{tr}$ and its retrieval performance is evaluated on $\mathcal{D}_{te}$.

\subsection{Tree Models}

Suppose $\mathcal{T}$ is a $b$-arity tree with height $H$, we regard the node at the $0$-th level as the root and nodes at the $H$-th level as leaves. 
Formally, we denote the node set at $h$-th level as $\mathcal{N}_h$ and the node set of $\mathcal{T}$ as $\mathcal{N}=\bigcup_{h=0}^H \mathcal{N}_h$.
For each node $n \in \mathcal{N}$, we denote its parent as $\rho (n) \in \mathcal{N}$, its children set as $\mathcal{C}(n) \subset \mathcal{N}$, the path from the root to it as $\mathrm{Path}(n)$, and the set of leaves on its subtree as $\mathcal{L}(n)$.

Tree models formulate the target set $\mathcal{I}$ as leaves of $\mathcal{T}$ through a bijective mapping $\pi: \mathcal{N}_H \to \mathcal{I}$, which implies $H=\mathrm{O}(\log_b M)$.
For any instance $(\xv,\yv)$, each node $n \in \mathcal{N}$ is defined with a pseudo target $z_n \in \{0,1\}$ to measure the existence of relevant targets on the subtree of $n$, i.e.,
\begin{equation}\label{eq:pseudotarget}
z_n = \mathbb{I}(\sum_{n' \in \mathcal{L}(n)} y_{\pi(n')} \geq 1),
\end{equation}
which satisfies $z_n = y_{\pi(n)}$ for $n \in \mathcal{N}_H$.

By doing so, tree models transform the original problem of estimating $p(y_j|\xv)$ to a series of hierarchical subproblems of estimating $p(z_n|\xv)$ on $n \in \mathrm{Path}(\pi^{-1}(j))$.
They introduce the node-wise scorer $g:\mathcal{X} \times \mathcal{N}\to \mathbb{R}$ to build such a node-wise estimator for each $n \in \mathcal{N}$, which is denoted as $p_g(z_n|\xv)$ to distinguish from the unknown distribution $p(z_n|\xv)$.
In the rest of this paper, we denote a tree model as $\mathcal{M}(\mathcal{T},g)$ to highlight its dependence on $\mathcal{T}$ and $g$.

\subsubsection{Training of Tree Models}

The training loss of tree models can be written as $\argmin_g \sum_{(\xv,\yv)\sim\mathcal{D}_{tr}} L(\yv,\gv(\xv))$, where
\begin{equation}\label{eq:treeloss}
	L(\yv,\gv(\xv)) =  \sum_{h=1}^H \sum_{n \in \mathcal{S}_h(\yv)} \ell_{\mathrm{BCE}} \left(z_n, g(\xv,n)\right).
\end{equation}
In Eq. (\ref{eq:treeloss}), $\gv(\xv)$ is a vectorized representation of $\{g(\xv,n): n\in \mathcal{N}\}$ (e.g., level-order traversal), $\ell_{\mathrm{BCE}}(z,g)=-z\log (1+\exp(-g))-(1-z)\log(1+\exp(g))$ is the binary cross entropy loss and $\mathcal{S}_h(\yv) \subset \mathcal{N}_h$ is the set of subsampled nodes at $h$-th level for an instance $(\xv,\yv)$.
Let $C=\max_h |\mathcal{S}_h(\yv)|$, the training complexity is $\mathrm{O}(HbC)$ per instance, which is logarithmic to the target set size $M$.

As two representatives of tree models, PLTs and TDMs adopt different ways\footnote{Details can be found in the supplementary materials.} to build $p_g$ and $S_h(\yv)$. 

{\bf PLTs:} Since $p(z_n|\xv)$ can be decomposed as $p(z_n=1|\xv)=\prod_{n' \in \mathrm{Path}(n)} p(z_{n'}=1|z_{\rho(n')}=1,\xv)$ according to Eq. (\ref{eq:pseudotarget}), $p_g(z_n|\xv)$ is decomposed accordingly via $p_g(z_{n'}|z_{\rho(n')}=1,\xv)=1/(1+\exp(-(2z_{n'}-1) g(\xv,n')))$.
As a result, only nodes with $z_{\rho(n)}=1$ are trained, which produces $\mathcal{S}_h(\yv)=\{n: z_{\rho(n)}=1, n \in \mathcal{N}_h\}$.

{\bf TDMs:} Unlike PLTs, $p(z_n|\xv)$ is estimated directly via $p_g(z_n|\xv)=1/(1+\exp(-(2z_n-1) g(\xv,n)))$. 
Besides, the subsample set\footnote{\citet{zhu2018learning} defines TDM with the constraint $|\mathcal{I}_{\xv}|=1$, we extend their definition by removing this constraint and refer TDM to such an extended definition in the rest of this paper.} is chosen as $\mathcal{S}_h(\yv)=\mathcal{S}^+_h(\yv) \bigcup \mathcal{S}^-_h(\yv)$ where $\mathcal{S}^+_h(\yv)=\{n: z_n=1, n \in \mathcal{N}_h\}$ and $\mathcal{S}^-_h(\yv)$ contains several random samples over $\mathcal{N}_h \setminus \mathcal{S}^+_h(\yv)$.

  
\subsubsection{Testing of Tree Models}  \label{sec:testing}
  
For any testing instance $(\xv,\yv)$, let $\mathcal{B}_h(\xv)$ denote the node set at $h$-th level retrieved by beam search and $k=|\mathcal{B}_h(\xv)|$ denote the beam size, the beam search process is defined as
\begin{equation}\label{eq:beam_gen}
	\mathcal{B}_h(\xv) \in \argTopk_{n \in \tilde{\mathcal{B}}_h(\xv)} p_g(z_n=1|\xv),
\end{equation} 
where $\tilde{\mathcal{B}}_h(\xv) = \bigcup_{n' \in \mathcal{B}_{h-1}(\xv)} \mathcal{C}(n')$.

By applying Eq. (\ref{eq:beam_gen}) recursively until $h=H$, beam search retrieves the set containing $k$ leaf nodes, denoted by $\mathcal{B}_H(\xv)$. Let $m \leq k$ denote the number of targets to be retrieved, the retrieved target subset can be denoted as 
\begin{equation}\label{eq:retrieve_target}
\hat{\mathcal{I}}_{\xv} = \{\pi(n): n \in \mathcal{B}_H^{(m)}(\xv)\},
\end{equation}
where $ \mathcal{B}_H^{(m)}(\xv) \in \argTopm_{n \in \mathcal{B}_H(\xv)} p_g(z_n=1|\xv)$ denote the subset of $\mathcal{B}_H(\xv)$ with top-$m$ scored nodes according to $p_g(z_n=1|\xv)$.
Since Eq. (\ref{eq:beam_gen}) only traverses at most $bk$ nodes and generating $\mathcal{B}_H(\xv)$ needs computing Eq. (\ref{eq:beam_gen}) for $H$ times, the testing complexity is $\mathrm{O}(Hbk)$ per instance, which is also logarithmic to $M$.

To evaluate the retrieval performance of $\mathcal{M}(\mathcal{T},g)$ on the testing dataset $D_{te}$, Precision@$m$, Recall@$m$ and F-measure@$m$ are widely adopted. 
Following \citet{zhu2018learning,zhu2019joint}, we define\footnote{Unlike macro/micro F-measure, the average of Eq. (\ref{eq:fmeasure}) over $D_{te}$ defines the instance-wise F-measure. \citet[Table 1]{wu2017unified} provides a thorough comparison for them.} them as the average of Eq. (\ref{eq:prec}), Eq. (\ref{eq:recall}) and Eq. (\ref{eq:fmeasure}) over $D_{te}$ respectively, where
\begin{equation}\label{eq:prec}
	\mathrm{P}@m (\mathcal{M};\xv,\yv) = \frac{1}{m}\sum\nolimits_{j \in \hat{\mathcal{I}}_{\xv}} y_j,
\end{equation}
\begin{equation}\label{eq:recall}
	\mathrm{R}@m (\mathcal{M};\xv,\yv) = \frac{1}{|\mathcal{I}_{\xv}|}\sum\nolimits_{j \in \hat{\mathcal{I}}_{\xv}} y_j,
\end{equation}
and
\begin{equation}\label{eq:fmeasure}
	\mathrm{F}@m (\mathcal{M};\xv,\yv) = \frac{2\cdot\mathrm{P}@m (\mathcal{M};\xv,\yv) 
\cdot \mathrm{R}@m (\mathcal{M};\xv,\yv)}{\mathrm{P}@m (\mathcal{M};\xv,\yv)+\mathrm{R}@m (\mathcal{M};\xv,\yv)}.
\end{equation}

\section{Main Contributions}\label{sec:main}

Our main contributions can be divided into three parts:
(1) We highlight the existence of the training-testing discrepancy on tree models, and provide an intuitive explanation of its negative effects on retrieval performance;
(2) We develop the concept of Bayes optimality under beam search and calibration under beam search to formalize this intuitive explanation;
(3) We propose a novel algorithm for learning tree models that are Bayes optimal under beam search.

\begin{figure*}[ht]
\vskip 0.2in
\begin{center}
\centerline{\includegraphics[width=\textwidth]{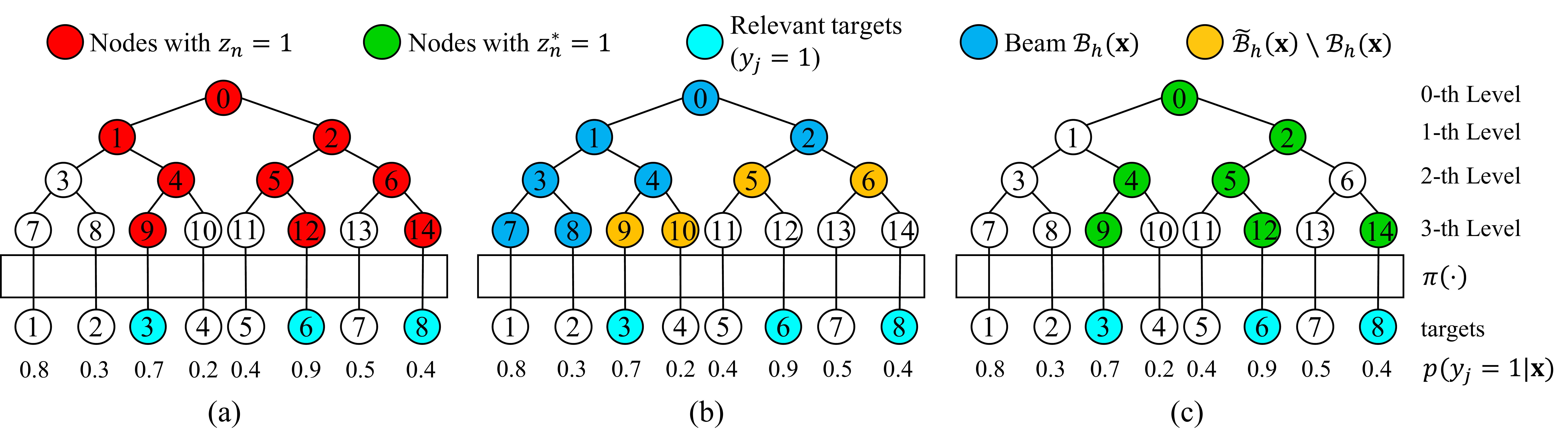}}
\vskip -0.1in
\caption{An overview of the training-testing discrepancy on a tree model $\mathcal{M}(\mathcal{T},g)$. (a) The assignment of pseudo targets on existing tree models, where red nodes correspond to $z_n=1$ defined in Eq. (\ref{eq:pseudotarget}). (b) Beam search process, where targets mapping blue nodes at 3-th level (i.e., leaf nodes) are regarded as the retrieval results of $\mathcal{M}$. (c) The assignment of optimal pseudo targets based on the ground truth distribution $\eta_j(\xv)=p(y_j=1|\xv)$, where green nodes correspond to $z_n^*=1$ defined in Eq. (\ref{eq:optimalpseudo}).}
\label{fig:optimalpseduo}
\end{center}
\vskip -0.26in
\end{figure*}

\subsection{Understanding the Training-Testing Discrepancy on Tree Models}\label{sec:drawback}

According to Eq. (\ref{eq:treeloss}), the training of $g(\xv,n)$ depends on two factors: the subsample set $\mathcal{S}_h(\yv)$ and the pseudo target $z_n$.
We can show that both factors relate to the training-testing discrepancy on existing tree models.

First, according to Eq. (\ref{eq:beam_gen}), the nodes at $h$-th level on which $g(\xv,n)$ is queried in testing can be denoted as $\tilde{\mathcal{B}}_h(\xv)$,
which implies a self-dependency of $g(\xv,n)$, i.e., nodes on which $g(\xv,n)$ is queried at $h$-th level depends on $g(\xv,n)$ queried at $(h-1)$-th level.
However, $\mathcal{S}_h(\yv)$, the nodes at $h$-th level on which $g(\xv,n)$ is trained, is generated according to ground truth targets $\yv$ via Eq. (\ref{eq:pseudotarget}).
Figure \ref{fig:optimalpseduo}(a) and Figure \ref{fig:optimalpseduo}(b) demonstrate such a difference: Node 7 and 8 (blue nodes) are traversed by beam search, but they are not in $\mathcal{S}_h(\yv)$ of PLTs and may not be in $\mathcal{S}_h(\yv)$ of TDMs according to $\mathcal{S}_h^+(\yv)$ (red nodes).
As a result, $g(\xv,n)$ is trained without considering such a self-dependency on itself when it is used for retrieving relevant targets via beam search.
This discrepancy results in that $g(\xv,n)$ trained well does not perform well in testing.

\begin{table}
\centering
\caption{Results for the toy experiment with $M=1000$, $b=2$. The reported number is $(\sum_{j \in \mathcal{I}^{(k)}} \eta_j - \sum_{j \in \hat{\mathcal{I}}}  \eta_j) / k$,
which is averaged over 100 runs with random initialization over $\mathcal{T}$ and $\eta_j$. 
}
\vskip 0.15in
\label{tab:toy}
\begin{tabular}{cccccc}
\toprule
$N$ & 100 & 1000 & 10000 & $\infty$ \\
\midrule
$k=1$ & 0.095 & 0.076 & 0.074 & 0.059 \\
$k=5$ & 0.075 & 0.055 & 0.050 & 0.037 \\
$k=10$ & 0.062 & 0.043 & 0.036 & 0.024 \\
$k=20$ & 0.057 & 0.036 & 0.031 & 0.018 \\
$k=50$ & 0.042 & 0.021 & 0.016 & 0.011 \\
\bottomrule
\end{tabular}
\vskip -0.12in
\end{table}

Second, $z_n$ defined in Eq. (\ref{eq:pseudotarget}) does not guarantee beam search w.r.t. $p_g(z_n=1|\xv)$ has no performance deterioration, i.e., retrieving the most relevant targets.
To see this, we design a toy example by ignoring $\xv$ and defining the data distribution to be $p(\yv)=\prod_{j=1}^M p(y_j)$, whose marginal probability $\eta_j=p(y_j=1)$ is sampled from a uniform distribution in $[0,1]$.
As a result, we denote the training dataset as $\mathcal{D}_{tr}=\{\yv^{(i)}\}_{i=1}^N$ and the pseudo target for instance $\yv^{(i)}$ on node $n$ as $z_n^{(i)}$.
For $\mathcal{M}(\mathcal{T},g)$, we assume $\mathcal{T}$ is randomly built and estimate $p(z_n=1)$ directly via\footnote{Estimating $p(z_n=1)$ hierarchically via $p_g(z_n=1)=\prod_{n' \in \mathrm{Path}(n)} p_g(z_{n'}=1|z_{\rho(n')}=1)$ and $p_g(z_{n}=1|z_{\rho(n)}=1)=\sum_{i=1}^N z_n^{(i)}z_{\rho(n)}^{(i)}/\sum_{i=1}^N z_{\rho(n)}^{(i)}$ provides similar results.} $p_g(z_n=1)=\sum_{i=1}^N z_n^{(i)}/N$ without the need to specify $g$, since there is no observation $\xv$.
Beam search with beam size $k$ is applied on $\mathcal{M}$ to retrieve the target subset whose size is $m=k$ as well, denoted by $\hat{\mathcal{I}}=\{\pi(n):n \in \mathcal{B}_H\}$.
Since $p(\yv)$ is known in this toy example, we need no testing set $\mathcal{D}_{te}$ and evaluate the retrieval performance directly via the regret $(\sum_{j \in \mathcal{I}^{(k)}} \eta_j - \sum_{j \in \hat{\mathcal{I}}}  \eta_j) / k$, where $\mathcal{I}^{(k)} \in \argTopk_{j \in \mathcal{I}} \eta_j$ denotes the top-$k$ targets according to $\eta_j$.
As a special case of Eq. (\ref{eq:regretbs}), this metric quantifies the suboptimality of $\mathcal{M}$ and we'll discuss it formally later.

As is shown in Table \ref{tab:toy}, we can find that the regret is always non-zero with varying training data number $N$ and beam size $k$.
Even in the ideal case when $N=\infty$ and thus $p_g(z_n=1)=p(z_n=1)$, it is still non-zero.
This implies that $z_n$ defined in Eq. (\ref{eq:pseudotarget}) cannot guarantee optimal retrieval performance in general.
This phenomenon does not contradict with the zero regret property in \citet[][Theorem 2]{wydmuch2018no}, since their theorem defines the regret using $\hat{\mathcal{I}}=\argTopk_{j \in \mathcal{I}} p_g(y_j=1|\xv)$, which ignores the performance deterioration caused by beam search.

\subsection{Bayes Optimality and Calibration under Beam Search}\label{sec:bayes}

In Sec. \ref{sec:drawback}, we discuss the existence of the training-testing discrepancy on tree models and provide a toy example to explain its effect.
Without loss of generality, we formalize this discussion with Precision@$m$ as the retrieval performance metric in this subsection.

The first question is, what does ``optimal'' mean for tree models with respect to their retrieval performance.
In fact, the answer has been partially revealed by the toy example in Sec. \ref{sec:drawback}, and we give a formal definition as follows:
\begin{definition}[Bayes Optimality under Beam Search]\label{def:bayesopt}
Given the beam size $k$ and the data distribution $p:\mathcal{X} \times \mathcal{Y} \to \mathbb{R}^+$, a tree model $\mathcal{M}(\mathcal{T},g)$ is called top-$k$ Bayes optimal under beam search if 
\begin{equation}\label{eq:bayesopt}
	\{\pi(n): n \in \mathcal{B}_H(\xv)\} \in \argTopk_{j \in \mathcal{I}} \eta_j(\xv),
\end{equation}
holds for any $\xv \in \mathcal{X}$.
$\mathcal{M}(\mathcal{T},g)$ is called Bayes optimal under beams search if Eq. (\ref{eq:bayesopt}) holds for any $\xv \in \mathcal{X}$ and $1\leq k \leq M$.
\end{definition}

Given Definition \ref{def:bayesopt}, we can derive a sufficient condition for
the existence of such an optimal tree model as follows\footnote{
Without any formal proof, \citet{zhu2018learning} proposes the max-heap like formulation, which can be regarded as a special case of Proposition \ref{thm:sufficient} with the $|\mathcal{I}_{\xv}|=1$ restriction.
We provide a detailed proof for Proposition \ref{thm:sufficient} in the supplementary materials.
}
:

\begin{prop}[Sufficient Condition for Bayes Optimality under Beam Search]\label{thm:sufficient}
Given the beam size $k$, the data distribution $p:\mathcal{X} \times \mathcal{Y} \to \mathbb{R}^+$, the tree $\mathcal{T}$ and  \begin{equation}\label{eq:optscorer}
	p^*(z_n|\xv)=\left\{
	\begin{aligned}
		&\max_{n' \in \mathcal{L}(n)} \eta_{\pi(n')}(\xv), & z_n=1 \\
		& 1-\max_{n' \in \mathcal{L}(n)} \eta_{\pi(n')}(\xv), & z_n=0 \\
	\end{aligned}
	\right. ,
\end{equation}
a tree model $\mathcal{M}(\mathcal{T},g)$ is top-$m$ Bayes optimal under beam search for any $m\leq k$, if $p_g(z_n|\xv) = p^*(z_n|\xv)$ holds for any $\xv \in \mathcal{X}$ and $n \in \bigcup_{h=1}^H \tilde{\mathcal{B}}_h(\xv)$.
$\mathcal{M}(\mathcal{T},g)$ is Bayes optimal under beam search, if $p_g(z_n|\xv) = p^*(z_n|\xv)$ holds for any $\xv \in \mathcal{X}$ and $n \in \mathcal{N}$.
\end{prop}

Proposition \ref{thm:sufficient} shows one case of what an optimal tree model should be, but it does not resolve all the problems, since both learning and evaluating a tree model require a quantitative measure of its suboptimality.
Notice that Eq. (\ref{eq:bayesopt}) implies that $\mathbb{E}_{p(\xv)}\left[\sum_{j \in \mathcal{I}_{\xv}^{(k)}} \eta_j(\xv) \right] = \mathbb{E}_{p(\xv)} \left[ \sum_{n \in \mathcal{B}_H(\xv)} \eta_{\pi(n)}(\xv)\right]$, where $\mathcal{I}_{\xv}^{(k)}=\argTopk_{j \in \mathcal{I}} \eta_j(\xv)$ denotes the top-$k$ targets according to the ground truth $\eta_j(\xv)$.
The deviation of such an equation can be used as a suboptimality measure of $\mathcal{M}$.
Formally, we define it to be the regret w.r.t. Precision@$k$ and denote it as $\mathrm{reg}_{p@k}(\mathcal{M})$.
This is a special case when $m=k$ for a more general definition $\mathrm{reg}_{p@m} (\mathcal{M}) = $
\begin{equation}\label{eq:regretbs}
	\mathbb{E}_{p(\xv)} \left[ \frac{1}{m} \left( \sum_{j \in \mathcal{I}_{\xv}^{(m)}} \eta_j(\xv) - \sum_{n \in \mathcal{B}_H^{(m)}(\xv)} \eta_{\pi(n)}(\xv) \right) \right],
\end{equation}
where $\mathcal{I}_{\xv}^{(m)}=\argTopm_{j \in \mathcal{I}} \eta_j(\xv)$.

Though $\mathrm{reg}_{p@k}(\mathcal{M})$ seems an ideal suboptimality measure, finding its minimizer is hard due to the existence of a series of nested non-differentiable $\argTopk$ operators.
Therefore, finding a surrogate loss for $\mathrm{reg}_{p@k}(\mathcal{M})$ such that its minimizer is still an optimal tree model becomes very important.
To distinguish such a surrogate loss, we introduce the concept of calibration under beam search as follows:

\begin{definition}[Calibration under Beam Search]\label{def:calibrate}
Given a tree model $\mathcal{M}(\mathcal{T},g)$, a loss function $L:\{0,1\}^M \times \mathbb{R}^{|\mathcal{N}|} \to \mathbb{R}$ is called top-$k$ calibrated under beam search if
\begin{equation}\label{eq:calibrate}
\begin{aligned}
 \argmin_g \mathbb{E}_{p(\xv,\yv)}\left[ L(\yv, \gv(\xv)) \right]   \subset \argmin_g \mathrm{reg}_{p@k}(\mathcal{M}),
\end{aligned}
\end{equation}
holds for any distribution $p:\mathcal{X} \times \mathcal{Y}\to \mathbb{R}^+$.
$L$ is called calibrated under beam search if Eq. (\ref{eq:calibrate}) holds for any $1\leq k \leq M$.
\end{definition}

Definition \ref{def:calibrate} shows a tree model $\mathcal{M}(\mathcal{T},g)$ with $g$ minimizing a non-calibrated loss is not Bayes optimal under beam search in general.
Recall that Proposition \ref{thm:sufficient} shows that for any $p:\mathcal{X} \times \mathcal{Y} \to \mathbb{R}^+$ and any $\mathcal{T}$, the minimizer of $\mathrm{reg}_{p@k}(\mathcal{M})$ always exists, which satisfies $p_g(z_n|\xv)=p^*(z_n|\xv)$ and achieves $\mathrm{reg}_{p@k}(\mathcal{M})=0$.
Therefore, the suboptimality of TDMs and PLTs can be proved by showing the minimizer of their training loss does not guarantee $\mathrm{reg}_{p@k}(\mathcal{M})=0$ in general.
This can be proved by finding a counterexample and the toy experiment shown in Table \ref{tab:toy} meets this requirement.
As a result, we have
\begin{prop}\label{thm:tdmplt}
Eq. (\ref{eq:treeloss}) with $z_n$ defined in Eq. (\ref{eq:pseudotarget}) is not calibrated under beam search in general.
\end{prop}

\subsection{Learning Optimal Tree Models under Beam Search}\label{sec:learning}

Given the discussion in Sec. \ref{sec:bayes}, we need a new surrogate loss function such that its minimizer corresponds to the tree model which is Bayes optimal under beam search.
According to Definition \ref{def:bayesopt}, when the retrieval performance is measured by Precision@$m$, requiring a model to be top-$m$ Bayes optimal under beam search will be enough.
Proposition \ref{thm:sufficient} provides a natural surrogate loss to achieve this purpose with beam size $k\geq m$, i.e.,
\begin{equation}\label{eq:calibratedloss}
g \in \argmin_g \mathbb{E}_{p(\xv)}\left[\sum_{h=1}^H \sum_{n \in \tilde{\mathcal{B}}_h(\xv)}\mathrm{KL}(p^*(z_n|\xv)\|p_g(z_n|\xv))\right],
\end{equation} 
where we follow the TDM style and assume $p_g(z_n|\xv)=1/(1+\exp(-(2z_n-1)g(\xv,n)))$.

Unlike Eq. (\ref{eq:treeloss}), Eq. (\ref{eq:calibratedloss}) uses nodes in $\tilde{\mathcal{B}}_h(\xv)$ instead of $\mathcal{S}^+_h(\yv)$ for training and introduces a different definition of pseudo targets compared to Eq. (\ref{eq:pseudotarget}).
Let $z_n^* \sim p^*(z_n|\xv)$ denote the corresponding pseudo target, we have
\begin{equation}\label{eq:optimalpseudo}
	z_n^* = y_{\pi(n')}, ~n' \in \argmax_{n'\in \mathcal{L}(n)} \eta_{\pi(n')}(\xv).
\end{equation}
Notice that for $n \in \mathcal{N}_H$, $z_n^*=y_{\pi(n)}$ as well as $z_n$ in Eq. (\ref{eq:pseudotarget}).
To distinguish $z_n^*$ from $z_n$, we call it the optimal pseudo target since it corresponds to the optimal tree model.
Given this definition, Eq. (\ref{eq:calibratedloss}) can be rewritten as $\argmin_g \mathbb{E}_{p(\xv,\yv)}\left[ L_p(\yv,\gv(\xv)) \right]$ where
\begin{equation}\label{eq:theoryloss}
	L_p(\yv,\gv(\xv))=\sum_{h=1}^H\sum_{n\in\tilde{\mathcal{B}}_h(\xv)} \ell_{\mathrm{BCE}}(z_n^*,g(\xv,n)).
\end{equation}

Notice that in Eq. (\ref{eq:theoryloss}) we assign a subscript $p$ to highlight the dependence of $z_n^*$ on $\eta_j(\xv)$, which implies that Eq. (\ref{eq:theoryloss}) is calibrated under beam search in the sense that its formulation depends on $p:\mathcal{X} \times \mathcal{Y} \to \mathbb{R}^+$.

Figure \ref{fig:optimalpseduo} provides a concrete example for the difference between $z_n^*$ and $z_n$.
Not all ancestor nodes of a relevant target $y_j=1$ are regarded as relevant nodes according to $z_n^*$:
Node $1$ and $6$ (red nodes in Figure \ref{fig:optimalpseduo}(a)) are assigned with $z_n=1$ but with $z_n^*=0$ (green nodes in Figure \ref{fig:optimalpseduo}(c)). 
The reason is that among targets on the subtree rooted at these nodes, the irrelevant target has a higher $\eta_j(\xv)$ compared to the relevant target, i.e., $\eta_7(\xv)=0.5>\eta_8(\xv)=0.4$ and $\eta_1(\xv)=0.8>\eta_3(\xv)=0.7$, which leads $z_n^*$ to be $0$. 

However, it is impossible to minimize Eq. (\ref{eq:theoryloss}) directly, since $\eta_j(\xv)$ is unknown in practice.
As a result, we need to find an approximation of $z_n^*$ without the dependence on $\eta_j(\xv)$.
Suppose $g(\xv,n)$ is parameterized with trainable parameters $\thetav \in \Thetav$,
we use the notation $g_{\thetav}(\xv,n)$, $p_{g_{\thetav}}(\xv)$ and ${\mathcal{B}_h} (\xv;\thetav)$ to highlight their dependence on $\thetav$.
A natural choice is to replace $\eta_{\pi(n')}(\xv)$ in Eq. (\ref{eq:optimalpseudo}) with $p_{g_{\thetav}}(z_{n'}=1|\xv)$.
However, this formulation is still impractical since the computational complexity of traversing $\mathcal{L}(n)$ for each $n \in \tilde{\mathcal{B}}_h(\xv;\thetav)$ is unacceptable. 
Thanks to the tree structure, we can approximate $z_n^*$ with $\hat{z}_n(\xv;\thetav)$, which is constructed in a recursive manner for $n \in \mathcal{N} \setminus \mathcal{N}_H$ as
\begin{equation}\label{eq:estoptimalpseudo}
	\hat{z}_n(\xv;\thetav) = \hat{z}_{n'}(\xv;\thetav), ~n' \in \argmax_{n'\in\mathcal{C}(n)} p_{g_{\thetav}}(z_{n'}=1|\xv),
\end{equation}
and is set directly as $\hat{z}_n(\xv;\thetav)=y_{\pi(n)}$ for $n \in \mathcal{N}_H$.

By doing so, we remove the dependence on unknown $\eta_j(\xv)$.
But minimizing Eq. (\ref{eq:theoryloss}) when replacing $z_n^*$ with $\hat{z}_n(\xv,\thetav)$ is still not an easy task since the parameter $\thetav$ affects $\tilde{\mathcal{B}}_h(\xv;\thetav)$, $\hat{z}_n(\xv,\thetav)$ and $g_{\thetav}(\xv,n)$: Gradient with respect to $\thetav$ cannot be computed directly due to the non-differentiability of the $\argTopk$ operator in $\tilde{\mathcal{B}}_h(\xv;\thetav)$ and the $\argmax$ operator in $\hat{z}_n(\xv;\thetav)$.
To get a differentiable loss function, we propose to replace $L_p(\yv,\gv(\xv))$ defined in Eq. (\ref{eq:theoryloss}) with 
\begin{equation}\label{eq:surrogateloss}
L_{\thetav_t}(\yv,\gv(\xv);\thetav) = \sum_{h=1}^H \sum_{n \in \tilde{\mathcal{B}}_h(\xv;\thetav_t)} \ell_{\mathrm{BCE}}(\hat{z}_n(\xv;\thetav_t), g_{\thetav}(\xv,n)),
\end{equation}
where $\thetav_t$ denotes the fixed parameter, which can be the parameter of the last iteration in a gradient based algorithm.
Given the discussion above, we propose a novel algorithm for learning such a tree model as Algorithm \ref{alg1}.

\begin{algorithm}
\begin{algorithmic}[1]
\caption{Learning Optimal Tree Models under Beam Search}
\label{alg1}
\REQUIRE Training dataset $\mathcal{D}_{tr}$, initial tree model $\mathcal{M}(\mathcal{T},g_{\thetav_0})$ with the tree structure $\mathcal{T}$ and the node-wise scorer $g_{\thetav_0}(\xv,n)$, beam size $k$, stepsize $\epsilon$.
\ENSURE Trained tree model $\mathcal{M}(\mathcal{T},g_{\thetav_t})$.
\STATE Initialize $t=0$;
\WHILE{convergence condition is not attained}
\STATE Draw a minibatch $\mathrm{MB}$ from $\mathcal{D}_{tr}$;
\STATE Draw $\tilde{\mathcal{B}}_h(\xv;\thetav_t)$ according to Eq. (\ref{eq:beam_gen});
\STATE Compute $\hat{z}_n(\xv;\thetav_t)$ for each $n\in \tilde{\mathcal{B}}_h(\xv;\thetav_t) $ according to Eq. (\ref{eq:estoptimalpseudo});
\STATE Update $\thetav_{t+1}$ using a gradient based method with stepsize $\epsilon$, e.g., ADAM \cite{kingma2014adam}, on the current $\thetav_t$ and the gradient $\gv_t$, where
$$
\gv_t = \nabla_{\thetav} \sum_{(\xv,\yv)\sim\mathrm{MB}} L_{\thetav_t}(\yv,\gv(\xv);\thetav)\Big|_{\thetav=\thetav_t}
$$
 according to Eq. (\ref{eq:surrogateloss});
\STATE $t \leftarrow t + 1$;
\ENDWHILE
\STATE Return $\mathcal{M}(\mathcal{T},g_{\thetav_t})$.
\end{algorithmic}
\end{algorithm}

As is analyzed in the supplementary materials, the training complexity of Algorithm \ref{alg1} is $\mathrm{O}(Hbk+Hb|\mathcal{I}_{\xv}|)$ per instance, which is still logarithmic to $M$. 
Besides, for the tree model trained according to Algorithm \ref{alg1}, its testing complexity is $\mathrm{O}(Hbk)$ per instance as that in Sec. \ref{sec:testing}, since Algorithm \ref{alg1} does not alter beam search in testing.

Now, the remaining question is, since introducing several approximations into Eq. (\ref{eq:surrogateloss}), does it still have the nice property to achieve Bayes optimality under beam search?
We provide an answer\footnote{Proof can be found in the supplementary materials.} as follows:
\begin{prop}[Practical Algorithm]\label{thm:practical}
Suppose $\mathcal{G} =\{g_{\thetav}:\thetav \in {\Thetav} \}$ has enough capacity and $L^*_{\thetav_t}(\yv,\gv(\xv);\thetav) =$
\begin{equation}\label{eq:starloss}
 \sum_{h=1}^H \sum_{n \in \mathcal{N}_h} w_n(\xv,\yv;\thetav_t) \ell_{\mathrm{BCE}}(\hat{z}_n(\xv;\thetav_t), g_{\thetav}(\xv,n)),
\end{equation}
where $w_n(\xv,\yv;\thetav_t) > 0$.
For any probability $p:\mathcal{X} \times \mathcal{Y} \to \mathbb{R}^+$, if there exists $\thetav_t\in \Thetav$ such that
\begin{equation}\label{eq:convergecond}
	\thetav_t \in \argmin_{\thetav \in \Thetav} \mathbb{E}_{p(\xv,\yv)}\left[ L^*_{\thetav_t}(\yv,\gv(\xv);\thetav) \right],
\end{equation}
the corresponding tree model $\mathcal{M}(\mathcal{T},g_{\thetav_t})$ is Bayes optimal under beam search.
\end{prop}

Proposition \ref{thm:practical} shows that replacing $z_n^*$ with $\hat{z}_n(\xv,\thetav)$ and introducing the fixed parameter $\thetav_t$ does not affect the optimality of $\mathcal{M}(\mathcal{T},g_{\thetav_t})$ on Eq. (\ref{eq:starloss}).
However, Eq. (\ref{eq:surrogateloss}) does not have such a guarantee, since the summation over $\tilde{\mathcal{B}}_h(\xv;\thetav_t)$ corresponds to the summation over $\mathcal{N}_h$ with weight $w_n(\xv,\yv;\thetav_t)=\mathbb{I}(n \in \tilde{\mathcal{B}}_h(\xv;\thetav_t))$ and thus violating the restriction that $w_n(\xv,\yv;\thetav_t)>0$.
This problem can be solved by introducing randomness into Eq. (\ref{eq:surrogateloss}) such that each $n \in \mathcal{N}_h$ has a non-zero $w_n(\xv,\yv;\thetav_t)$ in expectation.
Examples include adding random samples of $\mathcal{N}_h$ into the summation in Eq. (\ref{eq:surrogateloss}) or leveraging stochastic beam search \cite{kool2019stochastic} to generate $\tilde{\mathcal{B}}_h(\xv;\thetav_t)$.
Nevertheless, in experiments we find these strategies do not greatly affect the performance, and thus we still use Eq. (\ref{eq:surrogateloss}).

\section{Experiments}

In this section, we experimentally verify our analysis and evaluate the performance of different tree models on both synthetic and real data. 
Throughout experiments, we use OTM to denote the tree model trained according to Algorithm \ref{alg1} since its goal is to learn optimal tree models under beam search.
To perform an ablation study, we consider two variants of OTM:
OTM (-BS) differs from OTM by replacing $\tilde{\mathcal{B}}_h(\xv;\thetav_t)$ with $\mathcal{S}_h(\yv)=\mathcal{S}^+_h(\yv)\bigcup \mathcal{S}^-_h(\yv)$, and OTM (-OptEst) differs from OTM by replacing $\hat{z}_n(\xv;\thetav_t)$ in Eq. (\ref{eq:optimalpseudo}) with $z_n$ in Eq. (\ref{eq:pseudotarget}).
More details of experiments can be found in the supplementary materials.

\subsection{Synthetic Data}

{\bf Datasets:} For each instance $(\xv,\yv)$,
$\xv \in \mathbb{R}^d$ is sampled from a $d$-dimensional isotropic Gaussian distribution $\mathcal{N}(\zerov_d,\Iv_d)$ with zero mean and identity covariance matrix,
and $\yv \in \{0,1\}^M$ is sampled from $p(\yv|\xv)=\prod_{j=1}^M p(y_j|\xv)=\prod_{j=1}^M 1/(1+ \exp(-(2y_j-1)\wv_j^\top\xv-b))$ where the weight vector $\wv_j \in \mathbb{R}^d$ is also sampled from $\mathcal{N}(\zerov_d,\Iv_d)$. The bias $b$ is a predefined constant\footnote{In experiment, we set $b$ to be a negative value such that the number of non-zero entries is less than $0.1M$ to simulate the practical case where the number of relevant targets is much smaller than the target set size.} to control the number of non-zero entries in $\yv$.
Corresponding training and testing datasets are denoted as $\mathcal{D}_{tr}$ and $\mathcal{D}_{te}$, respectively.

{\bf Compared Models and Metric:} 
We compare OTM with PLT and TDM. 
All the tree models $\mathcal{M}(\mathcal{T},g)$ share the same tree structure $\mathcal{T}$ and the same parameterization of the node-wise scorer $g$.
More specifically, $\mathcal{T}$ is set to be a random binary tree over $\mathcal{I}$ and $g(\xv,n)=\thetav_n^{\top} \xv+b_n$ is parameterized as a linear scorer, where $\thetav_n \in \mathbb{R}^d$ and $b_n \in \mathbb{R}$ are trainable parameters.
All models are trained on $\mathcal{D}_{tr}$ and their perfomance is measured by $\widehat{\mathrm{reg}}_{p@m}$, which is an estimation of $\mathrm{reg}_{p@m}(\mathcal{M})$ defined in Eq. (\ref{eq:regretbs}) by replacing the expectation over $p(\xv)$ with the summation over $(\xv,\yv)\in \mathcal{D}_{te}$.

{\bf Results:} 
Table \ref{tab:synthetic_k} shows that OTM performs the best compared to other models, which indicates that eliminating the training-testing discrepancy can improve retrieval performance of tree models.
Both OTM (-BS) and OTM (-OptEst) have smaller regret than PLT and TDM, which means that using beam search aware subsampling (i.e., $\tilde{\mathcal{B}}_h(\xv;\thetav_t)$) or estimated optimal pseudo targets (i.e., $\hat{z}_n(\xv;\thetav_t)$) alone contributes to better performance.
Besides, OTM (-OptEst) has smaller regret than OTM (-BS), which reveals that beam search aware subsampling contributes more than estimated optimal pseudo targets to the performance of OTM.

\begin{table}
\centering
\caption{A comparison of $\widehat{\mathrm{reg}}_{p@m}(\mathcal{M})$ averaged by 5 runs with random initialization with hyperparameter settings $M=1000$, $d=10$, $b=-5$, $|\mathcal{D}_{tr}|=10000$, $|\mathcal{D}_{te}|=1000$ and $k=50$. }\label{tab:synthetic_k}
\vskip 0.15in
\begin{tabular}{ccccc}
\toprule
$m$ & $1$ & $10$ & $20$ & $50$ \\
\midrule
PLT & 0.0444 & 0.0778 & 0.0955 & 0.1492  \\
TDM & 0.0033 & 0.0205 & 0.0453 & 0.1363 \\
OTM & {\bf 0.0024} & {\bf 0.0163} & {\bf 0.0349} & {\bf 0.1083} \\
OTM (-BS) & 0.0048 & 0.0201 & 0.0421 & 0.1313 \\
OTM (-OptEst) & 0.0033 &  0.0198 & 0.0418 & 0.1218 \\
\bottomrule
\end{tabular}
\vskip -0.1in
\end{table}


\subsection{Real Data}

\begin{table*}
\centering
\caption{Precision@$m$, Recall@$m$ and F-Measure@$m$ comparison on Amazon Books with beam size $k=400$ and various $m$ (\%).}\label{tab:book}
\begin{tabular}{lcccccccccccc}
\toprule
\multirow{2}{4em}{Method} & \multicolumn{4}{c}{Precision} & \multicolumn{4}{c}{Recall} & \multicolumn{4}{c}{F-Measure}  \\
 & 10 & 50 & 100 & 200 & 10 & 50 & 100 & 200 & 10 & 50 & 100 & 200 \\
\midrule
Item-CF & 2.02 & 1.04 & 0.74 & 0.52 & 2.14 & 4.71 & 6.29 & 8.18 & 1.92 & 1.55 & 1.23 & 0.92 \\
YouTube product-DNN  & 1.26 & 0.84 & 0.67 & 0.53 & 1.12 & 3.52 & 5.41 & 8.26 & 1.05 & 1.21 & 1.09 & 0.93 \\
HSM & 1.50 & 0.93 & 0.73 & 0.54 & 1.25 & 3.59 & 5.59 & 8.04 & 1.21 & 1.30 & 1.18 & 0.95 \\
PLT & 1.85 & 1.26 &  0.99 & 0.75 & 1.57 & 4.87 & 7.35 & 10.59 & 1.48 & 1.74 & 1.57 & 1.29 \\
JTM & 1.84 & 1.34 & 1.07 & 0.80 & 1.75 & 5.79 & 8.70 & 12.60 & 1.60 & 1.94 & 1.73 & 1.40 \\ 
\midrule
OTM & {\bf 3.12} & {\bf 1.97} & {\bf 1.49} & {\bf 1.06} & {\bf 2.76} & {\bf 8.16} & {\bf 11.86} & {\bf 16.36} & {\bf 2.58} & {\bf 2.80} & {\bf 2.39} & {\bf 1.86} \\ 
OTM (-BS) & 2.18 & 1.45 & 1.15 & 0.86 & 1.91 & 6.01 & 9.40 & 13.68 & 1.81 & 2.08 & 1.88 & 1.52 \\
OTM (-OptEst)  & 3.07 & 1.92 & 1.45 & 1.05 & 2.70 & 8.00 & 11.63 & 16.17 & 2.54 & 2.74 & 2.33 & 1.83 \\ 
\bottomrule
\end{tabular}
\end{table*}

\begin{table*}
\centering
\caption{Precision@$m$, Recall@$m$ and F-Measure@$m$ comparison on UserBehavior with beam size $k=400$ and various $m$ (\%).}\label{tab:ub}
\begin{tabular}{lcccccccccccc}
\toprule
\multirow{2}{4em}{Method} & \multicolumn{4}{c}{Precision} & \multicolumn{4}{c}{Recall} & \multicolumn{4}{c}{F-Measure}  \\
 & 10 & 50 & 100 & 200 & 10 & 50 & 100 & 200 & 10 & 50 & 100 & 200 \\
\midrule
Item-CF & 5.45 & 3.07 & 2.20 & 1.56 & 1.25 & 3.31 & 4.74 & 6.75 & 1.84 & 2.76 & 2.64 & 2.30 \\
YouTube product-DNN & 9.04 & 4.52 & 3.22 & 2.25 & 2.29 & 5.36 & 7.49 & 10.15 & 3.29 & 4.23 & 3.97 & 3.36 \\
HSM & 9.79 & 4.49 & 3.04 & 2.01 & 2.58 & 5.60 & 7.38 & 9.52 & 3.68 & 4.30 & 3.80 & 3.03 \\
PLT & 11.47 & 5.07 & 3.47 & 2.35 & 2.85 & 5.84 & 7.75 & 10.22 & 4.13 & 4.72 & 4.23 & 3.48 \\
JTM & 20.05 & 7.45 & 4.85 & 3.12 & 5.42 & 9.39 & 11.84 & 14.75 & 7.62 & 7.15 & 6.06 & 4.70 \\
\midrule
OTM & {\bf 22.47} & {\bf 8.21} & {\bf 5.33} & {\bf 3.42} & {\bf 5.95} & {\bf 10.07} & {\bf 12.62} & {\bf 15.68} & {\bf 8.40} & {\bf 7.78} & {\bf 6.59} & {\bf 5.12} \\
OTM (-BS) & 19.81 & 7.74 & 5.08 & 3.31 & 5.36 & 9.57 & 12.14 & 15.29 & 7.54 & 7.36 & 6.30 & 4.95 \\
OTM (-OptEst) & 22.38 & 8.20 & {\bf 5.33} & 3.40 & 5.92 & 10.06 & 12.61 & 15.61 & 8.36 & {\bf 7.78} & {\bf 6.59} & 5.08 \\
\bottomrule
\end{tabular}
\end{table*}

{\bf Datasets:} Our experiment are conducted on two large-scale real datasets for recommendation tasks: Amazon Books \cite{mcauley2015image,he2016ups} and UserBehavior \cite{zhu2018learning}.
Each record of both datasets is organized in the format of user-item interaction, which contains user ID, item ID and timestamp.
The original interaction records are formulated as a set of user-based data.
Each user-based data is denoted as a list of items sorted by the timestep that the user-item interaction occurs.
We discard the user based data which has less than 10 items and split the rest into training set $\mathcal{D}_{tr}$, validation set $\mathcal{D}_{val}$ and testing set $\mathcal{D}_{te}$ in the same way as \citet{zhu2018learning,zhu2019joint}.
For the validation and testing set, we take the first half of each user-based data according to ascending order along timestamp as the feature $\xv$, and the latter half as the relevant targets $\yv$.
While training instances are generated from the raw user-based data considering the characteristics of different approaches on the training set. If the approach restricts $|\mathcal{I}_{\xv}|=1$, we use a sliding window to produce several instances for each user based data, while one instance is obtained for methods without restriction on $|\mathcal{I}_{\xv}|$.

{\bf Compared Models and Metric:} We compare OTM with two series of methods: (1) widely used methods in recommendation tasks, such as Item-CF \cite{sarwar2001item}, the basic collaborative filtering method, and YouTube product-DNN \cite{covington2016deep}, the representative work of vector kNN based methods; (2) tree models like HSM \cite{morin2005hierarchical}, PLT and JTM \cite{zhu2019joint}. HSM is a hierarchical softmax model which can be regarded as PLT with the $|\mathcal{I}_{\xv}|=1$ restriction.
JTM is a variant of TDM which trains tree structure and node-wise scorers jointly and achieves state-of-the-art performance on these two datasets. 
All the tree models share the same binary tree structure and adopt the same neural network model for node-wise scorers.
The neural network consists of three fully connected layers with hidden size 128, 64 and 24 and parametric ReLU is used as the activation function.  
The performance of different models is measured by Precision@$m$ (Eq. (\ref{eq:prec})), Recall@$m$ (Eq. (\ref{eq:recall})) and F-Measure@$m$ (Eq. (\ref{eq:fmeasure})) averaged over the testing set $\mathcal{D}_{te}$.

{\bf Results:} Table \ref{tab:book} and Table \ref{tab:ub} show results of Amazon Books and UserBehavior, respectively\footnote{As $k=400$, OTM is trained on $|\tilde{\mathcal{B}}_h(\xv;\thetav_t)|=800$ nodes per level.
For fairness in comparison, JTM also subsample $|\mathcal{S}_h(\yv)|=800$ nodes per level for training $g(\xv,n)$.}.
Our model performs the best among all methods:
Compared to the previous state-of-the-art JTM, OTM achieves $29.8\%$ and $6.3\%$ relative recall lift ($m=200$) on Amazon Books and UserBehavior separately.
Results of OTM and its two variants are consistent with that on synthetic data: Both beam search aware subsampling and estimated optimal pseudo targets contribute to better performance, while the former contributes more and the performance of OTM mainly depends on the former.
Besides, the comparison between HSM and PLT also demonstrates that removing the restriction of $|\mathcal{I}_{\xv}|=1$ in tree models contributes to performance improvement.

\begin{figure}
\begin{center}
\includegraphics[width=0.8\columnwidth]{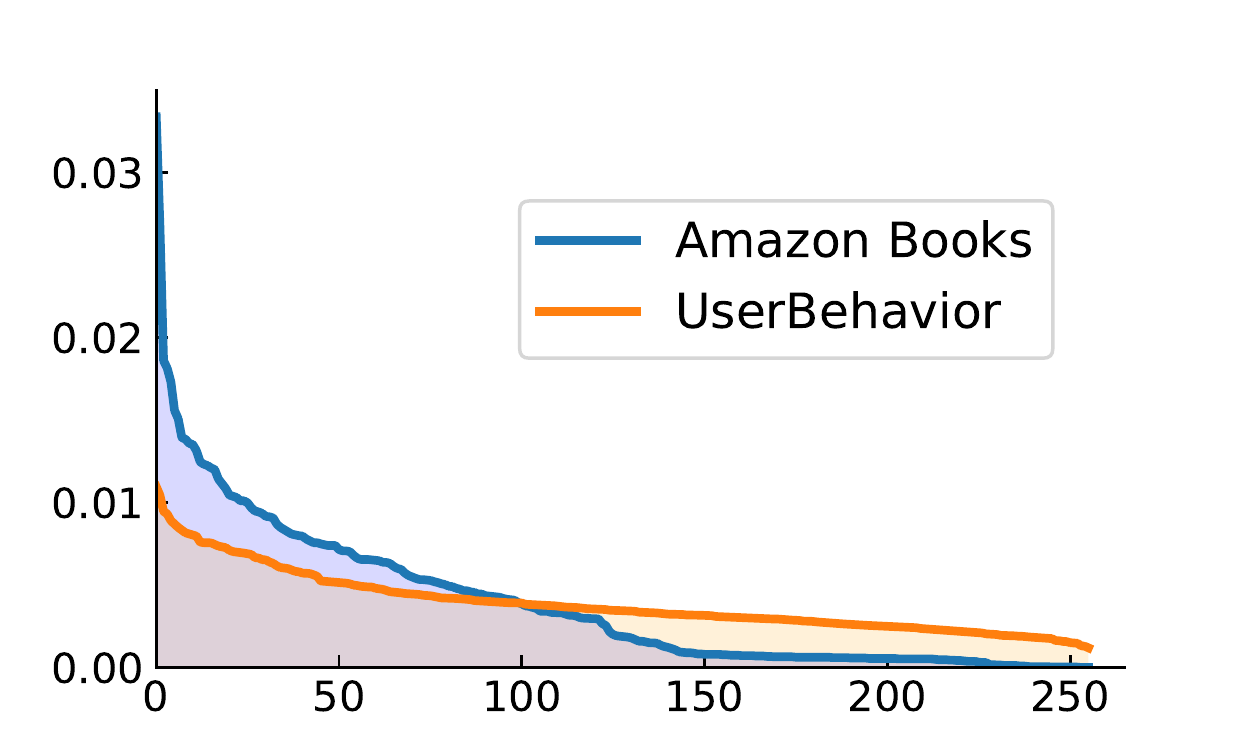}
\vskip -0.1in
\caption{Results of the level-wise distribution versus node index on Amazon Books and UserBehavior with $h=8$ ($|\mathcal{N}_h|=256$). }
\label{fig:explain}
\end{center}
\vskip -0.26in
\end{figure}

To understand why OTM achieves more significant improvement ($29.8\%$ versus $6.3\%$) on Amazon Books than UserBehavior, we analyze the statistics of these datasets and their corresponding tree structure.
For each $n \in \mathcal{N}$, we define $S_n = \sum_{(\xv,\yv)\in \mathcal{D}_{tr}} z_n$ to count the number of training instances which are relevant to $n$ (i.e., $z_n=1$).
For each level $1\leq h \leq H$, we sort $\{S_n:n\in \mathcal{N}_h\}$ in a descending order and normalize them as $S_n/\sum_{n'\in\mathcal{N}_h} S_{n'}$.
This produces a level-wise distribution, which reflects the data imbalance on relevant nodes resulted from the intrinsic property of both the datasets and the tree structure.
As is shown in Figure \ref{fig:explain}, the level-wise distribution of UserBehavior has a heavier tail than that of Amazon Books at the same level.
This implies the latter has a higher proportion of instances concentrated on only parts of nodes, which makes it easier for beam search to retrieve relevant nodes for training and thus leads to more significant improvement.

To verify our analysis on the time complexity of tree models, we compare their empirical training time, since they share the same beam search process in testing.
More specifically, we compute the wall-clock time per batch for training PLT, TDM and PLT with batch size 100 on the UserBehavior dataset. 
This number is averaged over 5000 training iterations on a single Tesla P100-PCIE-16GB GPU.
The results are $0.184$s for PLT, $0.332$s for TDM and $0.671$s for OTM, respectively.  
Though OTM costs longer time than PLT and JTM, they have the same order of magnitude. 
This is not weird, since the step 4 and 5 in Algorithm \ref{alg1} only increases the constant factor of complexity.
Besides, this is a reasonable trade-off for better performance and distributed training can alleviate this in practical applications.

\section{Conclusions and Future Work}

Tree models have been widely adopted in large-scale information retrieval and recommendation tasks due to their logarithmic computational complexity.
However, little attention has been paid to the training-testing discrepancy where the retrieval performance deterioration caused by beam search in testing is ignored in training.
To the best of our knowledge, we are the first to study this problem on tree models theoretically.
We also propose a novel training algorithm for learning optimal tree models under beam search which achieves improved experiment results compared to the state-of-the-arts on both synthetic and real data.

For future work, we'd like to explore other techniques for training $g(\xv,n)$ according to Eq. (\ref{eq:theoryloss}), e.g., the REINFORCE algorithm \cite{williams1992simple,ranzato2015sequence} and the actor-critic algorithm \cite{sutton2000policy,bahdanau2016actor}.
We also want to extend our algorithm for learning tree structure and node-wise scorers jointly.
Besides, applying our algorithm to applications like extreme multilabel text classification is also an interesting direction.

\section*{Acknowledgements}
We deeply appreciate Xiang Li, Rihan Chen, Daqing Chang, Pengye Zhang, Jie He and Xiaoqiang Zhu for their insightful suggestions and discussions. 
We thank Huimin Yi, Yang Zheng, Siran Yang, Guowang Zhang, Shuai Li, Yue Song and Di Zhang for implementing the key components of the training platform. 
We thank Linhao Wang, Yin Yang, Liming Duan and Guan Wang for necessary supports about online serving.
We thank anonymous reviewers for their constructive feedback and helpful comments.

\nocite{langley00}

\bibliography{bsat}
\bibliographystyle{icml2020}

\end{document}


\onecolumn

\icmltitle{Supplementary Material for Learning Optimal Tree Models under Beam Search}




\begin{icmlauthorlist}
\icmlauthor{Jingwei Zhuo}{to}
\icmlauthor{Ziru Xu}{to}
\icmlauthor{Wei Dai}{to}
\icmlauthor{Han Zhu}{to}
\icmlauthor{Han Li}{to}
\icmlauthor{Jian Xu}{to}
\icmlauthor{Kun Gai}{to}
\end{icmlauthorlist}

\icmlaffiliation{to}{Alibaba Group}
\icmlcorrespondingauthor{Jingwei Zhuo}{zjw169463@alibaba-inc.com}

\icmlkeywords{Machine Learning, ICML}

\vskip 0.3in

\appendix


\printAffiliationsAndNotice{}  


\setcounter{equation}{0}
\setcounter{table}{0}
\renewcommand\theequation{A.\arabic{equation}}
\renewcommand\thetable{A.\arabic{table}} 

This supplementary material consists of 3 sections: 
Sec. \ref{sec:a} introduces PLTs and TDMs in details and illustrates that both of them share the same training loss formulation;
Sec. \ref{sec:b} proves Proposition 1 and Proposition 3, and derives the computational complexity of Algorithm 1 in details; 
Sec. \ref{sec:c} gives the detailed settings and additional results of experiments.

\section{Detailed Introduction of PLTs and TDMs}\label{sec:a}

\subsection{Probabilistic Label Trees (PLTs)}

PLTs formulate tree models $\mathcal{M}(\mathcal{T},g)$ as hierarchical probability estimators for the marginal distribution $p(y_j|\xv)$ \cite{jain2016extreme,wydmuch2018no}.
In PLTs, the pseudo target $z_n$ is defined as $z_n = \mathbb{I}(\sum_{n' \in \mathcal{L}(n)} y_{\pi(n')} \geq 1)$, which implies that $z_n=1$ if and only if there exists $n' \in \mathcal{C}(n)$ such that $z_{n'}=1$.
In other words, $z_n=1$ implies $z_{\rho(n)}=1$.
As a result, for any $n \in \mathcal{N}$, corresponding $p(z_n|\xv)$ can be decomposed as
\begin{equation}\label{eq:derive1}
	 p(z_n=1|\xv) = \prod_{n' \in \mathrm{Path}(n)} p(z_{n'}=1|z_{\rho(n')}=1,\xv).
\end{equation}
Therefore, $p(y_j|\xv)$ can be represented as
\begin{equation}\label{eq:derive2}
	p(y_j|\xv) = p(z_{\pi^{-1}(j)}|\xv) =
\left\{
\begin{array}{ll}
\prod_{n \in \mathrm{Path}(\pi^{-1}(j))} p(z_{n}=1|z_{\rho(n)}=1,\xv), & y_j=1 \\
1-\prod_{n \in \mathrm{Path}(\pi^{-1}(j))} p(z_{n}=1|z_{\rho(n)}=1,\xv), & y_j=0 \\
\end{array}
\right.	
	 .
\end{equation}
According to Eq. (\ref{eq:derive2}), $\{p(y_j|\xv):j\in \mathcal{I}\}$ can be decomposed and represented by $\{p(z_n|z_{\rho(n)}=1,\xv):n \in \mathcal{N}\}$.
Leveraging this, PLTs transform the original probability estimation problem for $p(y_j|\xv)$ to a series of hierarchical estimation problems for $p(z_n|z_{\rho(n)}=1,\xv)$, whose corresponding probability estimator is formulated as $p_g(z_n|z_{\rho(n)}=1,\xv)=1/(1+\exp(- (2z_n-1) g(\xv,n)))$.
In other words, $g(\xv,n)$ is trained as a binary classifier for $z_n \sim p(z_n|z_{\rho(n)}=1,\xv)$.
Given an instance $(\xv,\yv)\in \mathcal{D}_{tr}$ where $\mathcal{D}_{tr}$ denotes the training dataset, $g(\xv,n)$ is trained only on the node $n \in \mathcal{N}$ whose parent satisfies $z_{\rho(n)} = 1$.
As a result, the loss function of PLTs can be denoted as $\sum_{(\xv,\yv)\sim \mathcal{D}_{tr}} L(\yv,\gv(\xv))$ with
\begin{equation}\label{eq:derive3}
\begin{aligned}
L(\yv,\gv(\xv)) & =\sum_{n \in \mathcal{N}} \mathbb{I}(z_{\rho(n)}=1) \ell_{\mathrm{BCE}}(z_n,g(\xv,n)) \\
& = \sum_{h=1}^H \sum_{n \in \mathcal{N}_h} \mathbb{I}(z_{\rho(n)}=1) \ell_{\mathrm{BCE}}(z_n,g(\xv,n)) \\
& = \sum_{h=1}^H \sum_{n \in \mathcal{C}(n')} \sum_{n' \in \mathcal{N}_{h-1}} \mathbb{I}(z_{n'}=1) \ell_{\mathrm{BCE}}(z_n,g(\xv,n)) \\
& = \sum_{h=1}^H \sum_{n \in \mathcal{S}_h(\yv)} \ell_{\mathrm{BCE}}(z_n,g(\xv,n)), \\
\end{aligned}
\end{equation}
where $\mathcal{S}_h(\yv)=\{n: n\in \mathcal{C}(n'), n'\in \mathcal{N}_{h-1}, z_{n'}=1\}=\{n:z_{\rho(n)}=1, n\in \mathcal{N}_h\}$.

\subsection{Tree-based Deep Models (TDMs)}

TDMs in the original paper \cite{zhu2018learning} only apply to the restricted case where $|\mathcal{I}_{\xv}|=1$, i.e., there exists only one target relevant to $\xv$.
For a training instance $(\xv,\yv)$ satisfying $\sum_{j \in \mathcal{I}}y_j=1$, let $j$ denote the single relevant target (i.e., $y_j=1$), TDMs assign each $n\in \mathcal{N}$ a pseudo target $z_n=1$ when $n$ is the ancestor node of $\pi^{-1}(j)$, and $z_n=0$ when $n$ is not the ancestor node of $\pi^{-1}(j)$.
They call $n \in \mathcal{N}$ a positive sample if $z_n=1$ and $n$ a negative sample if $z_n=0$.

According to the notations introduced in Sec. 2.1 of the main body, if $n$ is the ancestor node of $\pi^{-1}(j)$, the relationship between $n$ and $\pi^{-1}(j)$ can be represented as
\begin{equation}
	\pi^{-1}(j) \in \mathcal{L}(n) \Leftrightarrow \sum_{n' \in \mathcal{L}(n)} y_{\pi(n')} \geq 1,
\end{equation}
and thus the pseudo target $z_n$ defined in TDMs satisfies
\begin{equation}\label{eq:derive03}
	z_n = \mathbb{I}(\pi^{-1}(j) \in \mathcal{L}(n))  =  \mathbb{I}(\sum_{n' \in \mathcal{L}(n)} y_{\pi(n')} \geq 1),
\end{equation}
which coincides with the pseudo target definition of PLTs\footnote{In Eq. (\ref{eq:derive03}), under the restriction $\sum_{j \in \mathcal{I}}y_j=1$,  an equivalent representation of $\sum_{j\in \mathcal{I}} y_j \geq 1$ is $\sum_{j\in \mathcal{I}} y_j = 1$.}.

Unlike PLTs, TDMs formulate tree models $\mathcal{M}(\mathcal{T},g)$ to estimate $p(z_n|\xv)$ directly via $p_g(z_n|\xv)=1/(1+\exp(-(2z_n-1) g(\xv,n))$.
More specifically, $g(\xv,n)$ is formulated as a binary classifier for $z_n \sim p(z_n|\xv)$ instead of $z_n \sim p(z_n|z_{\rho(n)}=1,\xv)$.
To guarantee logarithmic computational complexity in training, TDMs leverage the idea of negative sampling, and constitute the subsample set of negative samples by randomly selecting nodes except the positive one at each level.
Let $\mathcal{S}_h^-(\yv)$ denote the subsample set of negative samples at $h$-th level and $\mathcal{S}_h^+(\yv)=\{n: z_n=1, n \in \mathcal{N}_h\}$ denote the set of the positive sample at $h$-th level, the training loss of TDMs can be denoted as $\sum_{(\xv,\yv)\sim \mathcal{D}_{tr}} L(\yv,\gv(\xv))$ where
\begin{equation}\label{eq:derive4}
\begin{aligned}
	L(\yv,\gv(\xv)) & = \sum_{h=1}^H \sum_{n \in \mathcal{S}_h^+(\yv) \bigcup  \mathcal{S}_h^-(\yv)} \ell_{\mathrm{BCE}}(z_n,g(\xv,n)) \\
	& = \sum_{h=1}^H \sum_{n \in \mathcal{S}_h(\yv)} \ell_{\mathrm{BCE}}(z_n,g(\xv,n)),
\end{aligned}
\end{equation}
where $\mathcal{S}_h(\yv) = \mathcal{S}_h^+(\yv) \bigcup  \mathcal{S}_h^-(\yv)$.

By far, we can find out that Eq. (\ref{eq:derive3}) and Eq. (\ref{eq:derive4}) follow the same formulation, which explains Eq. (2) in Sec. 3.	2.1 of the main body.
Besides, noticing that both Eq. (\ref{eq:derive03}) and Eq. (\ref{eq:derive4}) do not require $|\mathcal{I}_{\xv}|=1$, TDMs can be naturally extended to multiple relevant targets case by removing the original restriction $|\mathcal{I}_{\xv}|=1$ without changing its training loss formulation.

\section{Detailed Derivations}\label{sec:b}
\subsection{Proof of Proposition 1}
Given the condition in Proposition 1 that Eq. (9) holds for any $\xv \in \mathcal{X}$ and any $n\in \bigcup_{h=1}^H \tilde{\mathcal{B}}_h(\xv)$ with beam size $k$, our proof is divided into two parts: (1) proving $\mathcal{M}(\mathcal{T},g)$ is top-$m$ Bayes optimal under beam search when $m=k$; (2) proving $\mathcal{M}(\mathcal{T},g)$ is top-$m$ Bayes optimal under beam search for any $m < k$.
Notice that proving $\mathcal{M}(\mathcal{T},g)$ is Bayes optimal under beam search can be regarded as the beam size $k=M$ case of our proof, since $\tilde{\mathcal{B}}_h(\xv)=\mathcal{B}_h(\xv)=\mathcal{N}_h$ holds for any $1\leq h \leq H$ when $k=M$.
Besides, our proof does not rely on the assumption that there are no ties\footnote{This is a common assumption in previous papers, e.g., \citet{lapin2017analysis}. } among $\{\eta_j(\xv):j\in\mathcal{I}\}$ (that is, for any $i,j \in \mathcal{I}$ with $i \neq j$, $\eta_i(\xv)\neq \eta_j(\xv)$),
which makes our proof more general with the price that $\argmax$ and $\argTopm$ operators may have multiple solutions and thus the proof becomes more complex.

\subsubsection{Proof of The $m=k$ Case}\label{sec:proofmk}
Let $\mathcal{L}(\mathcal{B}_h(\xv))=\bigcup_{n \in \mathcal{B}_h(\xv)} \mathcal{L}(n)$ denote the leaf node set of every subtree rooted at $n \in \mathcal{B}_h(\xv)$, Eq. (8) in Definition 1, i.e., $\{\pi(n):n \in \mathcal{B}_H(\xv)\} \in \argTopk_{j\in\mathcal{I}} \eta_j(\xv)$, can be rewritten as
\begin{equation}\label{eq:proof100}
	\{\mathcal{B}_H(\xv)\} = \argTopk_{n \in \mathcal{B}_H(\xv)} \eta_{\pi(n)}(\xv) = \argTopk_{n \in \mathcal{L}(\mathcal{B}_H(\xv))} \eta_{\pi(n)}(\xv) \subset \argTopk_{n \in \mathcal{L}(\mathcal{B}_0(\xv))} \eta_{\pi(n)}(\xv) = \argTopk_{n \in \mathcal{N}_H} \eta_{\pi(n)}(\xv).
\end{equation}
In Eq. (\ref{eq:proof100}), the first equality holds since $|\mathcal{B}_H(\xv)|\leq k$ and thus $\mathcal{B}_H(\xv)$ is the unique solution of $\argTopk_{n \in \mathcal{B}_H(\xv)} \eta_{\pi(n)}(\xv)$, the second equality holds since $\mathcal{L}(\mathcal{B}_H(\xv)) = \mathcal{B}_H(\xv)$, the subset relationship holds since $\mathcal{B}_0(\xv)$ contains ancestor nodes of $\mathcal{B}_H(\xv)$ and thus $\mathcal{L}(\mathcal{B}_H(\xv)) \subset \mathcal{L}(\mathcal{B}_0(\xv))$, and the last equality holds since $\mathcal{B}_0(\xv)=\{r(\mathcal{T})\}$ and $\mathcal{L}(\{r(\mathcal{T})\}) = \mathcal{N}_H$.

As a result, Eq. (\ref{eq:proof100}) can be proved by showing that pruning $\tilde{\mathcal{B}}_h(\xv) \setminus \mathcal{B}_h(\xv)$ according to Eq. (3) does not lead to the retrieval performance deterioration where the top-$k$ targets w.r.t. $\eta_{\pi(n)}(\xv)$ among $ \mathcal{L}(\mathcal{B}_h(\xv))$ are also the top-$k$ targets w.r.t. $\eta_{\pi(n)}(\xv)$ among $\mathcal{N}_H$ (i.e., $\mathcal{I}$), i.e.,
\begin{equation}\label{eq:argtopkrelation0}
	\argTopk_{n \in \mathcal{L}(\mathcal{B}_H(\xv))} \eta_{\pi(n)}(\xv) \subset \cdots \subset \argTopk_{n \in \mathcal{L}(\mathcal{B}_{h}(\xv))} \eta_{\pi(n)}(\xv) \subset \cdots \subset \argTopk_{n \in \mathcal{L}(\mathcal{B}_0(\xv))} \eta_{\pi(n)}(\xv),
\end{equation}
which corresponds to showing
\begin{equation}\label{eq:argtopkrelation}
	\argTopk_{n \in \mathcal{L}(\mathcal{B}_h(\xv))} \eta_{\pi(n)}(\xv) \subset \argTopk_{n \in \mathcal{L}(\mathcal{B}_{h-1}(\xv))} \eta_{\pi(n)}(\xv),
\end{equation}
holds for any $\xv \in \mathcal{X}$ and any $1\leq h \leq H$. 

For $h$ such that $|\mathcal{N}_h| \leq k$, i.e., the number of nodes at $h$-th level is not larger than $k$, all nodes at $h$-th level are regarded as the beam, which corresponds to $\mathcal{B}_h(\xv)=\mathcal{N}_h$.
In this case, $|\mathcal{N}_{h-1}| \leq k$ also holds, which implies $\mathcal{L}(\mathcal{B}_h(\xv))=\mathcal{L}(\mathcal{N}_h)=\mathcal{L}(\mathcal{N}_{h-1})=\mathcal{L}(\mathcal{B}_{h=1}(\xv))$ and thus Eq. (\ref{eq:argtopkrelation}) always holds for any $\xv \in \mathcal{X}$.

For $h$ such that $|\mathcal{N}_h| > k$ and thus $|\mathcal{B}_h(\xv)|=k$, Eq. (\ref{eq:argtopkrelation}) can be proved by contradiction. Let $\mathcal{V}_h(\xv) \in \argTopk_{n \in \mathcal{L}(\mathcal{B}_h(\xv))} \eta_{\pi(n)}(\xv)$ denote the set of top-$k$ nodes among $\mathcal{L}(\mathcal{B}_h(\xv))$, 
we assume that there exists $\xv \in \mathcal{X}$ such that Eq. (\ref{eq:argtopkrelation}) does not holds, i.e.,  
\begin{equation}\label{eq:proofcontradassump}
\argTopk_{n \in \mathcal{L}(\mathcal{B}_h(\xv))} \eta_{\pi(n)}(\xv) \nsubseteq \argTopk_{n \in \mathcal{L}(\mathcal{B}_{h-1}(\xv))} \eta_{\pi(n)}(\xv).
\end{equation}
According to the definition of $\mathcal{L}(\mathcal{B}_h(\xv))$, $\mathcal{L}(\mathcal{B}_{h}(\xv)) \subset \mathcal{L}(\mathcal{B}_{h-1}(\xv))$, which indicates that both $\mathcal{V}_h(\xv)$ and $\mathcal{V}_{h-1}(\xv)$ are subsets of $\mathcal{L}(\mathcal{B}_{h-1}(\xv))$.
As a result, Eq. (\ref{eq:proofcontradassump}) holds if and only if
\begin{equation}\label{eq:proof1h000}
\sum_{n\in\mathcal{V}_{h-1}(\xv)} \eta_{\pi(n)}(\xv)>\sum_{n'\in\mathcal{V}_h(\xv)} \eta_{\pi(n')}(\xv).
\end{equation}
Eq. (\ref{eq:proof1h000}) implies that\footnote{Otherwise $\mathcal{V}_{h-1}(\xv) \subset \mathcal{L}(\mathcal{B}_h(\xv))$ which violates that $\mathcal{V}_h(\xv) \in \argTopk_{n \in \mathcal{L}(\mathcal{B}_h(\xv))} \eta_{\pi(n)}(\xv)$.} $\mathcal{V}_{h-1}(\xv) \setminus \mathcal{L}(\mathcal{B}_h(\xv)) \neq \emptyset$ and there exists $n_0 \in \mathcal{V}_{h-1}(\xv) \setminus \mathcal{L}(\mathcal{B}_h(\xv))$ such that
\begin{equation}\label{eq:needcontrad}
\eta_{\pi(n_0)}(\xv)>\min_{n'\in \mathcal{V}_h(\xv)} \eta_{\pi(n')}(\xv),
\end{equation}
otherwise $\sum_{n\in\mathcal{V}_h(\xv)} \eta_{\pi(n)}(\xv)=\sum_{n'\in\mathcal{V}_h'(\xv)} \eta_{\pi(n)}(\xv)$.

Let $\rho^{H-h}(n_0)=(\rho \circ \cdots \circ \rho) (n_0)$ denote the ancestor node of $n_0$ at $h$-th level, $n_0 \in \mathcal{V}_{h-1}(\xv) \setminus \mathcal{L}(\mathcal{B}_h(\xv))$ implies that  $\rho^{H-h}(n_0) \in \tilde{\mathcal{B}}_h(\xv) \setminus \mathcal{B}_h(\xv)$.
According to the definition of $\mathcal{B}_h(\xv)$ in Eq. (3), we have
\begin{equation}\label{eq:contrad00}
	p_g(z_{n}=1|\xv) \geq p_g(z_{\rho^{H-h}(n_0))}=1|\xv) \geq \eta_{\pi(n_0)}(\xv),~~\forall n \in \mathcal{B}_h(\xv).
\end{equation}
Recall that $p_g(z_{n}=1|\xv)=\max_{n' \in \mathcal{L}(n)}\eta_{\pi(n')}(\xv)$ holds for any $n \in \tilde{\mathcal{B}}_h(\xv)$ according to Eq. (9) of Proposition 1, Eq. (\ref{eq:contrad00}) can be rewritten as
\begin{equation}\label{eq:contrad11}
	\max_{n' \in \mathcal{L}(n)}\eta_{\pi(n')}(\xv) \geq \max_{n' \in \mathcal{L}(\rho^{H-h}(n_0)))}\eta_{\pi(n')}(\xv)\geq \eta_{\pi(n_0)}(\xv),~~\forall n \in \mathcal{B}_h(\xv), 
\end{equation}
where the last inequality holds since $n_0 \in \mathcal{L}(\rho^{H-h}(n_0))$.

According to Eq. (\ref{eq:contrad11}), for any $n \in \mathcal{B}_h(\xv)$, there exists at least one $n' \in \mathcal{L}(n)$ such that $\eta_{\pi(n')}(\xv) = \max_{n' \in \mathcal{L}(n)}\eta_{\pi(n')}(\xv) \geq \eta_{\pi(n_0)(\xv)}$.  
In other words, let $\mathcal{W}_h(\xv)=\{n': n' \in \argmax_{n' \in \mathcal{L}(n)}\eta_{\pi(n')}(\xv), n \in\mathcal{B}_h(\xv)\} \subset \mathcal{L}(\mathcal{B}_h(\xv))$ denote such nodes, we have $|\mathcal{W}_h(\xv)| \geq k$ and $\min_{n'\in \mathcal{W}_h(\xv)} \eta_{\pi(n')}(\xv) \geq \eta_{\pi(n_0)}(\xv)$.


Since $\mathcal{V}_h(\xv)$ denotes the top-$k$ nodes among $\mathcal{L}(\mathcal{B}_h(\xv))$, 
we have
\begin{equation}\label{eq:forcontrad}
\min_{n' \in \mathcal{V}_h(\xv)} \eta_{\pi(n')}(\xv) \geq \min_{n' \in \mathcal{W}_h(\xv)} \eta_{\pi(n')}(\xv) \geq \eta_{\pi(n_0)}(\xv).
\end{equation}
It is obvious that Eq. (\ref{eq:forcontrad}) contradicts with Eq. (\ref{eq:needcontrad}).
Therefore, the assumption does not hold and Eq. (\ref{eq:argtopkrelation}) always holds.
By doing so, we have proven that Eq. (\ref{eq:proof1h000}) holds for any $\xv \in \mathcal{X}$ and $1\leq h \leq H$, which indicates $\mathcal{M}(\mathcal{T},g)$ is top-$m$ Bayes optimal under beam search when $m=k$.

\subsubsection{Proof of The $m<k$ Case}

The $m<k$ case can be proved by reusing the proof of the $m=k$ case.
To see this, let $\mathcal{B}_h^{(m)}(\xv) \in \argTopm_{n \in \mathcal{B}_h(\xv)} p_g(z_n=1|\xv)$ denote the set of top-$m$ nodes w.r.t. $p_g(z_n=1|\xv)$ among $\mathcal{B}_h(\xv)$, which may not be the unique solution since we do not assume there exists no ties, we introduce a lemma as follows:
\begin{lemma}\label{thm:proof1}
Suppose that a tree model $\mathcal{M}(\mathcal{T},g)$ satisfies Eq. (9) for any $\xv \in \mathcal{X}$ and $n \in \bigcup_{h=1}^H \tilde{\mathcal{B}}_h(\xv)$ with beam size $k$.
For any $\xv \in \mathcal{X}$, $1 \leq h \leq H$, $1 \leq m \leq k$ and $\mathcal{B}_h^{(m)}(\xv)$, there always exists $\mathcal{B}_{h-1}^{(m)}(\xv)$ such that
\begin{equation}\label{eq:lemma}
	\mathcal{B}_h^{(m)}(\xv) \in \argTopm_{n \in \tilde{\mathcal{B}}_h^{(m)}(\xv)} p_g(z_n=1|\xv),~~\tilde{\mathcal{B}}_h^{(m)}(\xv) = \bigcup_{n' \in \mathcal{B}_{h-1}^{(m)}(\xv)} \mathcal{C}(n').
\end{equation}
\end{lemma}

\begin{proof}
	Eq. (\ref{eq:lemma}) follows the same formulation as Eq. (3) with the only difference in replacing $\mathcal{B}_h(\xv)$ with $\mathcal{B}_h^{(m)}$.
	
	For $h$ such that $|\mathcal{B}_h(\xv)| \leq m$, we have $\mathcal{B}_h^{(m)}(\xv)=\mathcal{B}_h(\xv)$ and thus there always exists $\mathcal{B}_{h-1}^{(m)}(\xv)=\mathcal{B}_{h-1}(\xv)$ such that Eq. (\ref{eq:lemma}) holds.
	
	For $h$ such that $|\mathcal{B}_h(\xv)| > m$, since Eq. (9) holds for any $n \in \bigcup_{h=1}^H \tilde{\mathcal{B}}_h(\xv)$, we have
	\begin{equation}\label{eq:lemma0}
		p_g(z_n=1|\xv) = \max_{n' \in \mathcal{L}(n)} \eta_{\pi(n')}(\xv) = \max_{n' \in \mathcal{C}(n)}\max_{n'' \in \mathcal{L}(n')} \eta_{\pi(n'')}(\xv) = \max_{n' \in \mathcal{C}(n)} p_g(z_{n'}=1|\xv),~~\forall n \in \bigcup_{h=1}^H \mathcal{B}_h(\xv).
	\end{equation}
	For any $\mathcal{B}_h^{(m)}(\xv) \in \argTopm_{n \in \mathcal{B}_h(\xv)} p_g(z_n=1|\xv)$, since $\mathcal{B}_h(\xv) \in \argTopk_{n \in \tilde{\mathcal{B}}_h(\xv)} p_g(z_n=1|\xv)$, we have $\mathcal{B}_h^{(m)}(\xv) \in \argTopm_{n \in \tilde{\mathcal{B}}_h(\xv)} p_g(z_n=1|\xv)$, i.e., $\mathcal{B}_h^{(m)}(\xv)$ is also the set of top-$m$ nodes among $\tilde{\mathcal{B}}_h(\xv)$, which is equivalent to
	\begin{equation}\label{eq:lemma01}
		\min_{n \in \mathcal{B}_{h}^{(m)}(\xv)} p_g(z_n=1|\xv) \geq \max _{n \in \tilde{\mathcal{B}}_h(\xv) \setminus \mathcal{B}_{h}^{(m)}(\xv)} p_g(z_{n}=1|\xv).
	\end{equation}
	As a result, let $\mathcal{A}_{h-1}(\xv) = \{\rho(n):n \in \mathcal{B}_h^{(m)}(\xv)\} \subset \mathcal{B}_{h-1}(\xv)$ denote the parent node set of $\mathcal{B}_h^{(m)}(\xv)$, 
	we have
	\begin{equation}\label{eq:lemma1}
	\begin{aligned}
		\min_{n \in \mathcal{A}_{h-1}(\xv)} p_g(z_n=1|\xv) & = \min_{n \in \mathcal{A}_{h-1}(\xv)}\max_{n' \in \mathcal{C}(n)} p_g(z_{n'}=1|\xv) \\
		& \geq \min_{n' \in \mathcal{B}_{h}^{(m)}(\xv)} p_g(z_{n'}=1|\xv) \\
		& \geq \max _{n' \in \tilde{\mathcal{B}}_h(\xv) \setminus \mathcal{B}_{h}^{(m)}(\xv)} p_g(z_{n'}=1|\xv) \\
		& \geq \max_{n \in \mathcal{B}_{h-1}(\xv) \setminus \mathcal{A}_{h-1}(\xv)} \max_{n' \in \mathcal{C}(n)} p_g(z_{n'}=1|\xv) \\
		& = \max_{n' \in \mathcal{B}_{h-1}(\xv) \setminus \mathcal{A}_{h-1}(\xv)} p_g(z_{n}=1|\xv).
	\end{aligned}
	\end{equation}
	In Eq. (\ref{eq:lemma1}), the first equality and the last equality holds because of Eq. (\ref{eq:lemma0}), the first inequality holds since\footnote{Otherwise there exists $n \in \mathcal{A}_{h-1}(\xv)$ such that for any $n' \in \argmax_{n' \in \mathcal{C}(n)} p_g(z_{n'}=1|\xv)$, $p_g(z_{n'}=1|\xv) > p_g(z_{n''}=1|\xv)$ always holds for any $n'' \in \mathcal{C}(n) \bigcap \mathcal{B}_h^{(m)}(\xv)$, which violates Eq. (\ref{eq:lemma01}).} $\argmax_{n' \in \mathcal{C}(n)} p_g(z_{n'}=1|\xv) \bigcap \mathcal{B}_h^{(m)}(\xv) \neq \emptyset$ always holds for any $n \in \mathcal{A}_{h-1}(\xv)$, the second inequality holds because of Eq. (\ref{eq:lemma01}), and the last inequality holds since $\{n':n' \in \mathcal{C}(n),~n \in \mathcal{B}_{h-1}(\xv) \setminus \mathcal{A}_{h-1}(\xv)\} \subset \tilde{\mathcal{B}}_h(\xv) \setminus \mathcal{B}_{h}^{(m)}(\xv)$.
	
	Now, suppose that $\mathcal{A}_h'(\xv)$ denotes the top-$(m-|\mathcal{A}_h(\xv)|)$ nodes among $\mathcal{B}_{h-1}(\xv) \setminus \mathcal{A}_{h-1}(\xv)$, i.e., 
	\begin{equation}
		\min_{n \in \mathcal{A}_{h-1}'(\xv)} p_g(z_n=1|\xv) \geq \max_{n' \in \mathcal{B}_{h-1}(\xv) \setminus \mathcal{A}_{h-1}(\xv) \setminus \mathcal{A}_{h-1}'(\xv)} p_g(z_{n'}=1|\xv),
	\end{equation}
	we have $|\mathcal{A}_{h-1}(\xv) \bigcup \mathcal{A}_{h-1}'(\xv)|=m$ and 
	\begin{equation}
		\min_{n \in \mathcal{A}_{h-1}(\xv) \bigcup \mathcal{A}_{h-1}'(\xv)} p_g(z_n=1|\xv) \geq \max_{n' \in \mathcal{B}_{h-1}(\xv) \setminus \mathcal{A}_{h-1}(\xv) \setminus \mathcal{A}_{h-1}'(\xv)} p_g(z_{n'}=1|\xv),
	\end{equation}
	which is equivalent to 
	\begin{equation}
		\mathcal{A}_{h-1}(\xv) \bigcup \mathcal{A}_{h-1}'(\xv) \in \argTopm_{n \in \mathcal{B}_{h-1}(\xv)} p_g(z_n=1|\xv).
	\end{equation}
	In other words, there always exists $\mathcal{B}_h^{(m)}(\xv)=\mathcal{A}_{h-1}(\xv) \bigcup \mathcal{A}_{h-1}'(\xv)$ such that Eq. (\ref{eq:lemma}) holds. 
	Therefore Lemma \ref{thm:proof1} has been proved.
\end{proof}

Lemma \ref{thm:proof1} indicates a nice property of $\mathcal{M}(\mathcal{T},g)$ in Proposition 1: The top-$m$ nodes among $\mathcal{B}_h(\xv)$, i.e., $\mathcal{B}_h^{(m)}(\xv)$, can be regarded as the generated beam of the top-$m$ nodes among $\mathcal{B}_{h-1}(\xv)$, i.e., $\mathcal{B}_{h-1}^{(m)}(\xv)$.
Besides, according to the definition of $\mathcal{B}_h^{(m)}(\xv)$, $\mathcal{B}_h^{(m)}(\xv) \subset \mathcal{B}_h(\xv)$ always holds, which implies that Eq. (9) also holds for any $n \in \bigcup_{h=1}^H \tilde{\mathcal{B}}_h^{(m)}(\xv)$ given the condition of Proposition 1.
Combining these two together, for any $\xv \in \mathcal{X}$, there exists $\{\mathcal{B}_h^{(m)}(\xv)\}_{h=1}^H$ satisfying Eq. (\ref{eq:lemma}) such that Eq. (9) also holds for any $n \in \bigcup_{h=1}^H \tilde{\mathcal{B}}_h^{(m)}(\xv)$.
Therefore, the top-$m$ Bayes optimality under beam search of $\mathcal{M}(\mathcal{T},g)$ for any $m<k$ can be proved by reusing the proof in Sec. \ref{sec:proofmk} to show that $\{\pi(n):n\in \mathcal{B}_H^{(m)}(\xv)\}\subset \argTopm_{j \in \mathcal{I}} \eta_j(\xv)$. 

Combining the proof of the $m=k$ case and the $m<k$ case, we have proven Proposition 1.

\subsection{Proof of Proposition 3}
Let $p_{g_{\thetav}}(z_n|\xv)=1/(1+\exp(-(2z_n-1) g_{\thetav}(\xv,n)))$, the loss function in Eq. (18) of Proposition 3 can be rewritten as
\begin{equation}\label{eq:proof3eq17}
\begin{aligned}
	&  \mathbb{E}_{p(\xv,\yv)}\left[ L_{\thetav_t}^*(\yv,\gv(\xv);\thetav)\right] \\
	= ~~~~&  \mathbb{E}_{p(\xv,\yv)}\left[ \sum_{h=1}^H \sum_{n \in \mathcal{N}_h}  w_n(\xv,\yv;\thetav_t)\Big(-\hat{z}_n(\xv;\thetav_t)\log p_{g_{\thetav}}(z_n=1|\xv)-(1-\hat{z}_n(\xv;\thetav_t))\log p_{g_{\thetav}}(z_n=0|\xv)\Big)\right] \\
	= ~~~~&  \mathbb{E}_{p(\xv)}\left[ \sum_{h=1}^H \sum_{n \in \mathcal{N}_h}  w_n(\xv,\yv;\thetav_t)\Big(-\mathbb{E}_{p(\yv|\xv)}\left[\hat{z}_n(\xv;\thetav_t)\right]\log p_{g_{\thetav}}(z_n=1|\xv)-(1-\mathbb{E}_{p(\yv|\xv)}\left[\hat{z}_n(\xv;\thetav_t)\right])\log p_{g_{\thetav}}(z_n=0|\xv)\Big)\right] \\
	= ~~~~& \mathbb{E}_{p(\xv)}\left[ \sum_{h=1}^H \sum_{n \in \mathcal{N}_h}  w_n(\xv,\yv;\thetav_t)\Big( \mathrm{KL}\left(\tilde{p}_{g_{\thetav_t}}(z_n|\xv)\| p_{g_{\thetav}}(z_n|\xv) \right) + \mathrm{H}(\tilde{p}_{g_{\thetav_t}}(z_n|\xv))\Big)\right], \\
\end{aligned}
\end{equation}
where 
\begin{equation}
\tilde{p}_{g_{\thetav_t}}(z_n|\xv)=
\left\{
\begin{array}{lr}
 \mathbb{E}_{p(\yv|\xv)}\left[\hat{z}_n(\xv;\thetav_t)\right], & z_n = 1 \\
 1- \mathbb{E}_{p(\yv|\xv)}\left[\hat{z}_n(\xv;\thetav_t)\right],& z_n = 0 \\ 
\end{array}
\right.
,
\end{equation}
and $\mathrm{H}(\tilde{p}_{g_{\thetav_t}}(z_n|\xv))=-\tilde{p}_{g_{\thetav_t}}(z_n=1|\xv) \log \tilde{p}_{g_{\thetav_t}}(z_n=1|\xv) -\tilde{p}_{g_{\thetav_t}}(z_n=0|\xv) \log \tilde{p}_{g_{\thetav_t}}(z_n=0|\xv)$ denotes the entropy of $p_{g_{\thetav_t}}(z_n|\xv)$.
Since $\mathrm{H}(\tilde{p}_{g_{\thetav_t}}(z_n|\xv))$ can be regarded as a constant term with respect to $\thetav$, Eq. (18) can be rewritten as
\begin{equation}\label{eq:deriveproof1}
	\thetav_t \in \argmin_{\thetav \in \Thetav} \mathbb{E}_{p(\xv)}\left[ \sum_{h=1}^H \sum_{n \in \mathcal{N}_h}  w_n(\xv,\yv;\thetav_t)\mathrm{KL}\left(\tilde{p}_{g_{\thetav_t}}(z_n|\xv)\| p_{g_{\thetav}}(z_n|\xv) \right)\right].
\end{equation}
According to the definition of KL divergence, the minimizer of $\mathrm{KL}\left(\tilde{p}_{g_{\thetav_t}}(z_n|\xv)\| p_{g_{\thetav}}(z_n|\xv) \right)$ satisfies $p_{g_{\thetav}}(z_n|\xv) = \tilde{p}_{g_{\thetav_t}}(z_n|\xv)$.
Since $\mathcal{G}$ has enough capacity (e.g., infinite capacity with non-parametric limit), such a minimizer can be obtained in $\Thetav$.
Therefore, if Eq. (\ref{eq:deriveproof1}) holds, we have $p_{g_{\thetav}}(z_n|\xv) = \tilde{p}_{g_{\thetav_t}}(z_n|\xv)$ when $\thetav=\thetav_t$ for any $\xv\in\mathcal{X}$ and $n\in\mathcal{N}$.

For any $n \in \mathcal{N} \setminus \mathcal{N}_H$, we have
\begin{equation}\label{eq:proof3iter}
\begin{aligned}
	p_{g_{\thetav_t}}(z_n=1|\xv) & = \tilde{p}_{g_{\thetav_t}}(z_n=1|\xv) \\
	& =\mathbb{E}_{p(\yv|\xv)}\left[\hat{z}_n(\xv;\thetav_t)\right] \\
	& = \mathbb{E}_{p(\yv|\xv)}\left[\hat{z}_{n'}(\xv;\thetav_t)\right],~n'\in\argmax_{n'\in\mathcal{C}(n)} p_{g_{\thetav_t}}(z_{n'}=1|\xv) \\
	& = \tilde{p}_{g_{\thetav_t}}(z_{n'}=1|\xv),~n'\in\argmax_{n'\in\mathcal{C}(n)} p_{g_{\thetav_t}}(z_{n'}=1|\xv) \\	
	& = p_{g_{\thetav_t}}(z_{n'}=1|\xv),~n'\in\argmax_{n'\in\mathcal{C}(n)} p_{g_{\thetav_t}}(z_{n'}=1|\xv) \\	
	& = \max_{n'\in \mathcal{C}(n)} p_{g_{\thetav_t}}(z_{n'}=1|\xv),~\forall \xv \in \mathcal{X}.
\end{aligned}
\end{equation}
For any $n \in \mathcal{N}_H$, $\hat{z}_n(\xv;\thetav_t)=y_{\pi(n)}$ according to Eq. (15), and thus we have
\begin{equation}\label{eq:proof3h0}
	p_{g_{\thetav_t}}(z_n=1|\xv) = \tilde{p}_{g_{\thetav_t}}(z_n=1|\xv)= \mathbb{E}_{p(\yv|\xv)}\left[ y_{\pi(n)} \right] = \eta_{\pi(n)}(\xv), ~\forall \xv \in \mathcal{X}.
\end{equation}

Now, assuming that $p_{g_{\thetav_t}}(z_n=1|\xv)=\max_{n'\in \mathcal{L}(n)}\eta_{\pi(n')}(\xv)$ holds for any $n \in \mathcal{N}_{h+1}$, for any $n \in \mathcal{N}_h$, we have
\begin{equation}
\begin{aligned}
p_{g_{\thetav_t}}(z_n=1|\xv) & = \max_{n'\in \mathcal{C}(n)} p_{g_{\thetav_t}}(z_{n'}=1|\xv) \\
	& = \max_{n'\in \mathcal{C}(n)} \max_{n'' \in \mathcal{L}(n')} \eta_{\pi(n'')}(\xv) \\ 
	& = \max_{n'' \in \mathcal{L}(n)} \eta_{\pi(n'')(\xv)}.
\end{aligned}
\end{equation}
By doing so, we have proven that $p_{g_{\thetav_t}}(z_n=1|\xv)= \max_{n'\in\mathcal{L}(n)}\eta_{\pi(n')}(\xv)$ holds for any $\xv \in \mathcal{X}$ and $n \in \mathcal{N}$, i.e., $p_{g_{\thetav_t}}(z_n|\xv)= \tilde{p}(z_n|\xv)$ holds for any $\xv \in \mathcal{X}$ and $n \in \mathcal{N}$.
According to Proposition 1, we can conclude that $\mathcal{M}(\mathcal{T},g_{\thetav_t})$ is Bayes optimal under beam search.

\subsection{Computational Complexity of Algorithm 1}

In Algorithm 1, the computational complexity depends on step 4 and step 5, where the former retrieves nodes according to beam search and the latter estimates the optimal pseudo target for each node retrieved by beam search.
Recall that $(\xv,\yv)$ denotes an instance and $\yv \in \{0,1\}^M$ is an equivalent representation of $\mathcal{I}_{\xv}$ (the relevant target subset), we analyze the complexity per instance as follows.

For step 4, there are at most $b k$ nodes needed to be queried at each level according to Eq. (3) and the tree has $H$ levels.
Therefore, its complexity is $\mathrm{O}(Hbk)$.

For step 5, according to the definition in Eq. (15), $\hat{z}_n(x;\theta)=0$ always holds if $\mathcal{L}(n) \bigcap \mathcal{I}_{\xv} = \emptyset$, i.e., $n \notin \mathcal{S}_h^+(\yv)$ for any $h \in \{1,...,H\}$.
Therefore, at the $h$-th level, we only need to compute $\hat{z}_n(\xv;\thetav)$ for $n \in \tilde{\mathcal{B}}_h(\xv;\thetav) \bigcap \mathcal{S}_h^+(\yv)$ and set $\hat{z}_n(\xv;\thetav)=0$ directly for $n \in \tilde{\mathcal{B}}_h(\xv;\thetav) \setminus \mathcal{S}_h^+(\yv)$.
In the worst case, we need to compute $\hat{z}_n(\xv;\thetav)$ for each $n \in \mathcal{S}_h^+(\yv)$ at the $h$-th level, which can be computed recursively in a bottom-up manner.
According to Eq. (15), computing $\hat{z}_n(\xv;\thetav)$ needs to query the children set $\mathcal{C}(n)$ of $n \in \mathcal{S}_h^+(\yv)$, where $\hat{z}_{n'}(\xv;\thetav)$ for $n' \in \mathcal{C}(n) \bigcap \mathcal{S}_{h+1}^+(\yv)$ has been computed while that for $n' \in \mathcal{C}(n) \setminus \mathcal{S}_{h+1}^+(\yv)$ is always zero.
As a result, it needs to query $|\mathcal{C}(n)| \leq b$ nodes.
Since $|\mathcal{S}_h^+(\yv)|\leq |\mathcal{I}_{\xv}|$ and the tree has $H$ levels, the complexity of step 5 is $\mathrm{O}(Hb|\mathcal{I}_{\xv}|)$.

To conclude, the computational complexity of Algorithm 1 is $\mathrm{O}(Hbk+Hb|\mathcal{I}_{\xv}|)$.

\section{Experiments}\label{sec:c}

\subsection{Toy Example}

\begin{table*}
\centering
\caption{Results for the toy experiment with $M=1000$, $b=2$. The reported number is $\mathrm{reg}_{p@m}(\mathcal{M})$ under different hyperparameter settings of $m$, $k$ and $N$, 
and is averaged over 100 runs with random initialization over $\mathcal{T}$ and $\eta_j$.}\label{tab:toyfull}
\vskip 0.1in
\begin{tabular}{cccccccccccccc}
\toprule
\multirow{2}{0em}{k} & \multirow{2}{0em}{m} &  \multicolumn{4}{c}{DirEst} & \multicolumn{4}{c}{HierEst} & \multicolumn{4}{c}{OptEst}  \\
& & 100 & 1000 & 10000 & $\infty$ & 100 & 1000 & 10000 & $\infty$ & 100 & 1000 & 10000 & $\infty$ \\
\midrule
1 & 1 & 0.088 & 0.083 & 0.079 & 0.059 & 0.093 & 0.078 & 0.081 & 0.059 & 0.009 & 0.000 & 0.000 & 0.000 \\
\midrule
5 & 1 & 0.023 & 0.013 & 0.012 & 0.007 & 0.021 & 0.012 & 0.011 & 0.007 & 0.009 & 0.000 & 0.000 & 0.000 \\
5 & 5 & 0.073 & 0.052 & 0.044 & 0.032 & 0.071 & 0.051 & 0.046 & 0.032 & 0.007 & 0.000 & 0.000 & 0.000 \\
\midrule
10 & 1 & 0.014 & 0.006 & 0.004 & 0.002 & 0.014 & 0.006 & 0.005 & 0.002 & 0.008 & 0.001 & 0.000 & 0.000 \\
10 & 5 & 0.031 & 0.019 & 0.015 & 0.008 & 0.031 & 0.018 & 0.016 & 0.008 & 0.007 & 0.000 & 0.000 & 0.000 \\
10 & 10 & 0.064 & 0.046 & 0.039 & 0.023 & 0.063 & 0.045 & 0.039 & 0.023 & 0.005 & 0.001 & 0.000 & 0.000 \\
\midrule
20 & 1 & 0.010 & 0.003 & 0.002 & 0.001 & 0.011 & 0.003 & 0.002 & 0.001 & 0.009 & 0.001 & 0.000 & 0.000 \\
20 & 5 & 0.017 & 0.008 & 0.006 & 0.002 & 0.017 & 0.008 & 0.006 & 0.002 & 0.007 & 0.000 & 0.000 & 0.000 \\ 
20 & 10 & 0.028 & 0.015 & 0.013 & 0.006 & 0.028 & 0.016 & 0.013 & 0.006 & 0.005 & 0.001 & 0.000 & 0.000 \\
20 & 20 & 0.059 & 0.038 & 0.033 & 0.020 & 0.060 & 0.039 & 0.033 & 0.020 & 0.005 & 0.000 & 0.000 & 0.000 \\
\midrule
50 & 1 & 0.009 & 0.001 & 0.000 & 0.000 & 0.009 & 0.001 & 0.000 & 0.000 & 0.009 & 0.001 & 0.000 & 0.000 \\
50 & 5 & 0.008 & 0.001 & 0.001 & 0.000 & 0.009 & 0.002 & 0.001 & 0.000 & 0.007 & 0.000 & 0.000 & 0.000 \\
50 & 10 & 0.011 & 0.002 & 0.001 & 0.000 & 0.011 & 0.003 & 0.001 & 0.000 & 0.005 & 0.001 & 0.000 & 0.000 \\
50 & 20 & 0.017 & 0.005 & 0.003 & 0.001 & 0.017 & 0.005 & 0.003 & 0.001 & 0.005 & 0.000 & 0.000 & 0.000 \\
50 & 50 & 0.042 & 0.021 & 0.016 & 0.011 & 0.042 & 0.021 & 0.016 & 0.011 & 0.005 & 0.001 & 0.000 & 0.000 \\ 
\bottomrule
\end{tabular}
\vskip -0.1in
\end{table*}

The toy example in Sec. 4.1 investigates the retrieval performance of a tree model $\mathcal{M}(\mathcal{T},g)$ whose pseudo targets are defined in Eq. (1).
Given the training dataset $\mathcal{D}_{tr}=\{\yv^{(i)}\}_{i=1}^N$, $\mathcal{M}(\mathcal{T},g)$ is trained to estimate the node-wise probability of $z_n$ directly via $p_g(z_n=1)=\sum_{i=1}^N z_n^{(i)}/N$, where $z_n^{(i)}=\mathbb{I}(\sum_{n'\in\mathcal{L}(n)} y_{\pi(n)}^{(i)} \geq 1)$.
Table 1 shows $\mathcal{M}(\mathcal{T},g)$ with such $p_g(z_n=1)$ have non-zero regret in general, which corresponds to the retrieval performance deterioration.

In this subsection, we provide additional experimental results for this toy example.
More specifically, we consider three different methods for building $p_g(z_n=1)$, i.e.,
\begin{itemize}
	\item Direct Estimator (DirEst)\footnote{The method used in Table 1 corresponds to DirEst.}: $p_g(z_n=1)=\sum_{i=1}^N z_n^{(i)}/N$, where $z_n^{(i)}=\mathbb{I}(\sum_{n'\in \mathcal{L}(n)} y_{\pi(n')}^{(i)} \geq 1)$ (i.e., Eq. (1));
	\item Hierarchical Estimator (HierEst): $p_g(z_n=1)=\prod_{n' \in \mathrm{Path}(n)} p_g(z_{n'}=1|z_{\rho(n')}=1)$ with $p_g(z_{n}=1|z_{\rho(n)}=1)=\sum_{i=1}^N z_n^{(i)}z_{\rho(n)}^{(i)}/\sum_{i=1}^N z_{\rho(n)}^{(i)}$, where $z_n^{(i)}=\mathbb{I}(\sum_{n'\in \mathcal{L}(n)} y_{\pi(n')}^{(i)} \geq 1)$ and $z_{\rho(n)}^{(i)}=\mathbb{I}(\sum_{n'\in \mathcal{L}(\rho(n))} y_{\pi(n')}^{(i)} \geq 1)$ (i.e., Eq. (1));
	\item Optimal Estimator (OptEst): $p_g(z_n=1)=\sum_{i=1}^N z_n^{(i)}/N$, where $z_n=z_{n'}$ with $n' \in \argmax_{n'\in\mathcal{C}(n)} p_g(z_{n'}=1)$ (i.e., Eq. (15)). 
\end{itemize}

We use the abbreviation DirEst, HierEst and OptEst to denote these three methods.
Table \ref{tab:toyfull} shows corresponding experimental results.
Comparing DirEst and HierEst, we can find that both of them produce similar results for any choices of $k$, $m$ and $N$, which verifies the rationality of only providing results of DirEst in Table 1.
Besides, $\mathrm{reg}_{p@m}(\mathcal{M})$ of both DirEst and HierEst is non-zero in general\footnote{$\mathrm{reg}_{p@m}(\mathcal{M})$ seems to be zero for cases when $N=\infty$, $k=50$ and $m=1$, $5$ or $10$. This is because the reported number is rounded to three decimal places.}, which is consistent with the results in Table 1.
Compared to both DirEst and HierEst, OptEst achieves much smaller $\mathrm{reg}_{p@m}(\mathcal{M})$ for any choices of $k$, $m$ and $N$. 
In the ideal case when $N=\infty$, OptEst achieves zero regret.
These findings verify the correctness of the optimal pseudo target definition in Eq. (13) and the rationality of its recursive estimation in Eq. (15).
For these three methods, a common phenomenon is that given the same $m$ and $N$, increasing $k$ leads to smaller $\mathrm{reg}_{p@m}(\mathcal{M})$, i.e., better retrieval performance. 
The reason is that a larger beam size $k$ corresponds to a larger leaf level beam set $\mathcal{B}_H$ on which the $\argTopm$ operator is used to retrieve targets\footnote{An extreme case is that $k =M$ and $N=\infty$ such that $\mathcal{B}_H=\mathcal{N}_H$, $p_g(z_n=1)=\eta_{\pi(n)}$, respectively.
In this case, no matter which $p_g(z_n=1)$ is used, $\mathrm{reg}_{p@m}(\mathcal{M})$ always equals zero.}.



\subsection{Synthetic Data}

Recall that in Sec. 5.1, we set $c=-5$ to simulate the practical case when the number of relevant targets is much smaller than the target set size. 
In other words, such a $c$ can be regarded as a sparsity controller for the target set.
In this subsection, we provide a thorough comparison of PLT, TDM and OTM on the synthetic data with various sparsity of relevant targets.
More specifically, we choose $c$ in $\{0,-1,-2,-3,-4,-5\}$, which results in that the ratio of relevant targets per instance becomes $50.07\%, 38.57\%, 28.26\%,  19.38\%,  12.92\%, 8.14\%$, respectively.

\begin{table}
\centering
\caption{A comparison of $\widehat{\mathrm{reg}}_{p@m}(\mathcal{M})$ averaged by 5 runs with random initialization under hyperparameter settings $M=1000$, $d=10$, $|\mathcal{D}_{tr}|=10000$, $|\mathcal{D}_{te}|=1000$, $k=50$ and various $c$. }\label{tab:synthetic_k}
\subtable[$c=0$.]{
\centering
\begin{tabular}{ccccc}
\toprule
 $m$ & $1$ & $10$ & $20$ & $50$ \\
\midrule
 PLT & 0.0008 & 0.0024 & 0.0041 & 0.0150  \\
 TDM & 0.0005 & 0.0021 & 0.0037 & 0.0148 \\
 OTM & 0.0006 & 0.0015 & 0.0026 & {\bf 0.0088} \\
 OTM (-BS) & {\bf 0.0001} & {\bf 0.0007} & {\bf 0.0019} & 0.0110 \\
 OTM (-OptEst) & 0.0007 & 0.0025 & 0.0048 & 0.0139 \\
\bottomrule
\end{tabular}
}
\subtable[$c=-1$.]{
\begin{tabular}{ccccc}
\toprule
 $m$ & $1$ & $10$ & $20$ & $50$ \\
\midrule
 PLT & 0.0022 & 0.0056 & 0.0091 & 0.0302  \\
 TDM & 0.0004 & 0.0024 & 0.0057 & 0.0280 \\
 OTM & 0.0006 & 0.0021 & 0.0044 & {\bf 0.0182} \\
 OTM (-BS) & {\bf 0.0002} & {\bf 0.0016} & {\bf 0.0042} & 0.0242 \\
 OTM (-OptEst) & 0.0005  & 0.0023 & 0.0054 & 0.0262 \\
\bottomrule
\end{tabular}
}
\subtable[$c=-2$.]{
\begin{tabular}{ccccc}
\toprule
 $m$ & $1$ & $10$ & $20$ & $50$ \\
\midrule
 PLT & 0.0055 & 0.0124 & 0.0189 & 0.0564  \\
 TDM & 0.0008 & 0.0045 & 0.0111 & 0.0538 \\
 OTM & 0.0008 & 0.0036 & {\bf 0.0082} & {\bf 0.0369} \\
 OTM (-BS) & {\bf 0.0004} & {\bf 0.0033} & 0.0091 & 0.0502 \\
 OTM (-OptEst) & 0.0007 & 0.0039 & 0.0097 & 0.0481 \\
\bottomrule
\end{tabular}
}
\subtable[$c=-3$.]{
\begin{tabular}{ccccc}
\toprule
 $m$ & $1$ & $10$ & $20$ & $50$ \\
\midrule
 PLT & 0.0145 &  0.0278 & 0.0394 & 0.0978  \\
 TDM & 0.0013 & 0.0087 & 0.0214 & 0.0919  \\
 OTM & {\bf 0.0008} & {\bf 0.0061} & {\bf 0.0149} & {\bf 0.0648} \\
 OTM (-BS) & 0.0008 & 0.0070 & 0.0183 & 0.0878 \\
 OTM (-OptEst) & 0.0011 & 0.0076 & 0.0187 & 0.0826 \\
\bottomrule
\end{tabular}
}
\subtable[$c=-4$.]{
\begin{tabular}{ccccc}
\toprule
 $m$ & $1$ & $10$ & $20$ & $50$ \\
\midrule
 PLT & 0.0281 & 0.0542 & 0.0693 & 0.1335  \\
 TDM & 0.0022 & 0.0139 & 0.0319 & 0.1224 \\
 OTM & {\bf 0.0012} & {\bf 0.0093} & {\bf 0.0225} & {\bf 0.0906} \\
 OTM (-BS) & 0.0012 & 0.0106 & 0.0278 & 0.1183 \\
 OTM (-OptEst) & 0.0014 & 0.0116 & 0.0278 & 0.1090 \\
\bottomrule
\end{tabular}
}
\subtable[$c=-5$.]{
\begin{tabular}{ccccc}
\toprule
 $m$ & $1$ & $10$ & $20$ & $50$ \\
\midrule
 PLT & 0.0444 & 0.0778 & 0.0955 & 0.1492  \\
 TDM & 0.0033 & 0.0205 & 0.0453 & 0.1363 \\
 OTM & {\bf 0.0024} & {\bf 0.0163} & {\bf 0.0349} & {\bf 0.1083} \\
 OTM (-BS) & 0.0048 & 0.0201 & 0.0421 & 0.1313 \\
 OTM (-OptEst) & 0.0033 &  0.0198 & 0.0418 & 0.1218 \\
\bottomrule
\end{tabular}
}
\vskip -0.2in
\end{table}

Results are shown in Table \ref{tab:synthetic_k}.
We can find that OTM and its variants perform better than PLT and TDM in general.
An interesting phenomenon is the different behaviors of the beam search aware subsampling (BS) and the estimating optimal pseudo targets (OptEst) on various $c$.
OTM (-BS) has lower $\widehat{\mathrm{reg}}_{p@m}(\mathcal{M})$ compared to PLT and TDM for all choices of $c$, which reflects that OptEst contributes to better performance consistently.
By comparing the results between OTM (-OptEst) and OTM (-BS), we can find that OTM (-OptEst) performs worse than OTM (-BS) when $c$ is large (e.g., $c=0$) and OTM (-OptEst) performs better than OTM (-BS) when $c$ is small (e.g., $c=-5$), which implies that BS plays different roles according to the sparsity of relevant targets: When relevant targets are sparse, BS contributes to better performance; When relevant targets are not sparse, BS negatively affect retrieval performance.
Besides, another interesting phenomenon is that OTM performs the best when $m=k$, while for the $m<k$ case, OTM may perform worse than OTM (-BS).
Since $k=50$ is fixed, the retrieved beam $\mathcal{B}_H(\xv)$ is also fixed and the performance for varying $m$ only depends on $\{g(\xv,n): n\in \mathcal{B}_H(\xv)\}$.
Therefore, such a phenomenon implies that $g(\xv,n)$ on the leaf level may not be well trained to preserve the order information among $\{\eta_{\pi(n)}(\xv):n\in \mathcal{B}_H(\xv)\}$.
We'd like to analyze this phenomenon in details for future work.


\subsection{Real Data}

\begin{table}
\centering
\caption{Statistics of Amazon Books and UserBehavior datasets.}\label{tab:realstat}
\vskip 0.1in
\begin{tabular}{ccc}
\toprule
& Amazon Books & UserBehavior \\
\midrule
Num. of users & 294,739 & 969,529 \\
Num. of items & 1,477,922 & 4,162,024 \\
Num. of records & 8,654,619 & 100,020,395 \\
\bottomrule
\end{tabular}
\vskip -0.1in
\end{table}

\begin{table*}
\centering
\caption{Precision@$m$, Recall@$m$ and F-Measure@$m$ comparison between \citet{zhu2019joint} and our implementation, with beam size $k=400$ and $m=200$. The percent sign ($\%$) is omitted for each number.}\label{tab:ub}
\vskip 0.1in
\begin{tabular}{llcccccc}
\toprule
\multirow{2}{4em}{Method} & & \multicolumn{3}{c}{Amazon Books} & \multicolumn{3}{c}{UserBehavior}   \\
 & & Precision & Recall & F-Measure & Precision & Recall & F-Measure \\
\midrule
\multirow{2}{4em}{HSM} & \citet{zhu2019joint} & 0.42 & 6.22 & 0.72 & 1.80 & 8.62 & 2.71 \\
& Our implementation & 0.54 &  8.04 & 0.95 & 2.01 & 9.52 & 3.03 \\
\midrule
\multirow{2}{4em}{JTM} &\citet{zhu2019joint}  & 0.79 & 12.45 & 1.38 & 3.11 & 14.71 & 4.68 \\
& Our implementation & 0.80 & 12.60 & 1.40 & 3.12 & 14.75 & 4.70 \\
\bottomrule
\end{tabular}
\vskip -0.1in
\end{table*}

Table \ref{tab:realstat} summarizes the statistics of Amazon Books and UserBehavior datasets.

{\bf Preprocessing:} In the main body we briefly introduce how to preprocess these two datasets.

{\bf Implementation:} 
The node-wise scorer $g(\xv,n)$ is built as follows:
After preprocessing, for each instance $(\xv,\yv)$, $\xv \in \mathcal{I}^{69}$ is a 69-dimensional vector where $x_t$ ($1\leq t\leq 69$) denotes the user's behavior (i.e., the interacted item) at $t$-th nearest time w.r.t. the earliest interaction time of item in $\yv \in \{0,1\}^{|\mathcal{I}_{\xv}|}$.
For any $n\in \mathcal{N}_h$, $\xv$ is transformed\footnote{This is called the hierarchical user preference representation, which is proposed in \citet{zhu2019joint}.} to a level-wise representation $\xv (n)$ by replacing $x_t$ with $x_t(n)=\rho^{H-h}(\pi^{-1}(x_t))$.
Both $x_t(n)$ and $n$ are embeded to be 24-dimensional continuous vectors denoted by $\mathrm{emb}(x_t(n))$ and $\mathrm{emb}(n)$, respectively.
According to the time order from near to far, $\{\mathrm{emb}(x_t(n))\}_{t=1}^{69}$ is further split into 10 windows with window size 1, 1, 1, 2, 2, 2, 10, 10, 20, 20, respectively. 
Then, average pooling is applied to produce a 24-dimensional vector for each window.
Finally, these 10 vectors are concatenated with the node embedding vector and produces a 264-dimensional vector as the input to the following neural network, 
which consists of three fully-connected layers, with 128, 64 and 24 hidden units
and Parametric ReLU as the activation function. 
The tree hierarchy $\mathcal{T}$ is chosen to be the one produced by JTM\footnote{Recall that JTM optimizes $\mathcal{T}$ and $g(\xv,n)$ jointlty.} and is shared by all the tree models, including HSM, PLT, JTM and OTM. 
By doing so, all the tree models have the same tree hierarchy and the same formulation of node-wise scorers, 
the difference of retrieval performance of these models can be attributed to the difference of training algorithms of them,
and thus different tree models can be compared fairly under this setting.

{\bf Results:}
In Table 3 and Table 4 of the main body, 
the results for YouTube product-DNN and HSM are produced using codes in \nolinkurl{https://github.com/alibaba/x-deeplearning/tree/master/xdl-algorithm-solution/TDM/} and the results for JTM are produced using codes provided by the supplemental of \nolinkurl{https://papers.nips.cc/paper/8652-joint-optimization-of-tree-based-index-and-deep-model-for-recommender-systems}.

We also provide additional experimental results for comparing the retrieval performance reported in the original JTM paper \cite{zhu2019joint} and that in our implementation, which is shown in Table \ref{tab:ub}.
For JTM, we can find that our implementation achieves similar results to that reported in the original paper, which verifies the rationality of our implementation.
An interesting observation is that our implemented HSM achieves better results compared to the original one.
The reason is that in our experiment settings, HSM uses the same $\mathcal{T}$ and $g(\xv,n)$ formulation as JTM. 
While in the original paper, HSM uses a different $\mathcal{T}$ which is built according to category information of raw dataset and $g(\xv,n)$ does not use the hierarchical user preference representation.

\nocite{langley00}

\bibliography{bsat}
\bibliographystyle{icml2020}